\documentclass[journal]{IEEEtran}
\usepackage{microtype}
\usepackage{enumitem}
\usepackage{graphicx}
\usepackage{subcaption}
\usepackage{booktabs}
\usepackage{siunitx}
\usepackage{algorithm, algorithmic}
\usepackage[colorlinks, linkcolor=black, anchorcolor=black, citecolor=black]{hyperref}
\usepackage{stfloats}
\usepackage{amsmath}
\usepackage{amssymb}
\usepackage{mathtools}
\usepackage{amsthm}
\usepackage{bm}
\usepackage{multirow}
\usepackage{makecell}
\usepackage{caption}
\usepackage{graphicx}
\usepackage{float}
\usepackage[capitalize,noabbrev]{cleveref}
\usepackage{url}
\usepackage{xcolor}

\usepackage[
    per-mode = symbol,
]{siunitx}
\usepackage{fontawesome5}
\newcommand{\orcidicon}[1]{%
    \href{https://orcid.org/#1}{\textcolor[HTML]{A6CE39}{\faOrcid}}%
}
\begin{document}

\title{Fault-Tolerant MARL for CAVs under Observation Perturbations for Highway On-Ramp Merging}

\author{Yuchen Shi\textsuperscript{\orcidicon{0009-0006-1021-8478}}, Huaxin Pei\textsuperscript{\orcidicon{0000-0003-4815-2778}}, Yi Zhang\textsuperscript{\orcidicon{0000-0001-5526-866X}},~\IEEEmembership{Senior Member,~IEEE}, Danya Yao\textsuperscript{\orcidicon{0000-0001-5032-6322}},~\IEEEmembership{Member,~IEEE}

\thanks{This work was supported in part by the National Natural Science Foundation of China under Grant 62503259. (\emph{Corresponding author: Huaxin Pei.})}
\thanks{Yuchen Shi and Huaxin Pei are with the Department of Automation, Tsinghua University, Beijing 100084, China (e-mail: shiyuche21@mails.tsinghua.edu.cn; phx17@tsinghua.org.cn).}%
\thanks{Yi Zhang and Danya Yao are with the Department of Automation, Beijing National Research Center for Information Science and Technology (BNRist), Tsinghua University, Beijing 100084, China. (e-mail: zhyi@tsinghua.edu.cn; yaody@tsinghua.edu.cn).}%
}%

\maketitle

\begin{abstract}
    Multi-Agent Reinforcement Learning (MARL) holds significant promise for enabling cooperative driving among Connected and Automated Vehicles (CAVs). However, its practical application is hindered by a critical limitation, i.e., insufficient fault tolerance against observational faults. Such faults, which appear as perturbations in the vehicles' perceived data, can substantially compromise the performance of MARL-based driving systems. Addressing this problem presents two primary challenges. One is to generate adversarial perturbations that effectively stress the policy during training, and the other is to equip vehicles with the capability to mitigate the impact of corrupted observations. To overcome the challenges, we propose a fault-tolerant MARL method for cooperative on-ramp vehicles incorporating two key agents. First, an adversarial fault injection agent is co-trained to generate perturbations that actively challenge and harden the vehicle policies. Second, we design a novel fault-tolerant vehicle agent equipped with a self-diagnosis capability, which leverages the inherent spatio-temporal correlations in vehicle state sequences to detect faults and reconstruct credible observations, thereby shielding the policy from misleading inputs. Experiments in a simulated highway merging scenario demonstrate that our method significantly outperforms baseline MARL approaches, achieving near-fault-free levels of safety and efficiency under various observation fault patterns.
\end{abstract}


\section{Introduction}

\IEEEPARstart{M}{ulti-vehicle} cooperative autonomous driving is anticipated to significantly enhance traffic efficiency and safety \cite{Xu2019Group, Pei2021Cooperative, 9569746, Yu_Yang_Zhong_Yang_Fan_Luo_Nie_2025}. Among the driving scenarios, highway on-ramp merging represents a critical bottleneck where coordination faults can lead to severe congestion and safety hazards. Through real-time information exchange, CAVs can acquire a more comprehensive understanding of traffic conditions, enabling them to make more efficient and safer decisions. However, unavoidable communication and perception faults \cite{domajnko2024hardware, 7827613, Wang2025Malicious} cause the observed data to deviate from the ground truth. Such observation perturbations directly mislead the decision-making of vehicles receiving faulty data, subsequently propagating the disruptive impact throughout the collaborative system. This propagation can cause the vehicle fleet to lose its cooperative decision-making capability and potentially trigger safety incidents \cite{feng2018vulnerability}.

\begin{figure*}[htbp]
\includegraphics[width=\linewidth]{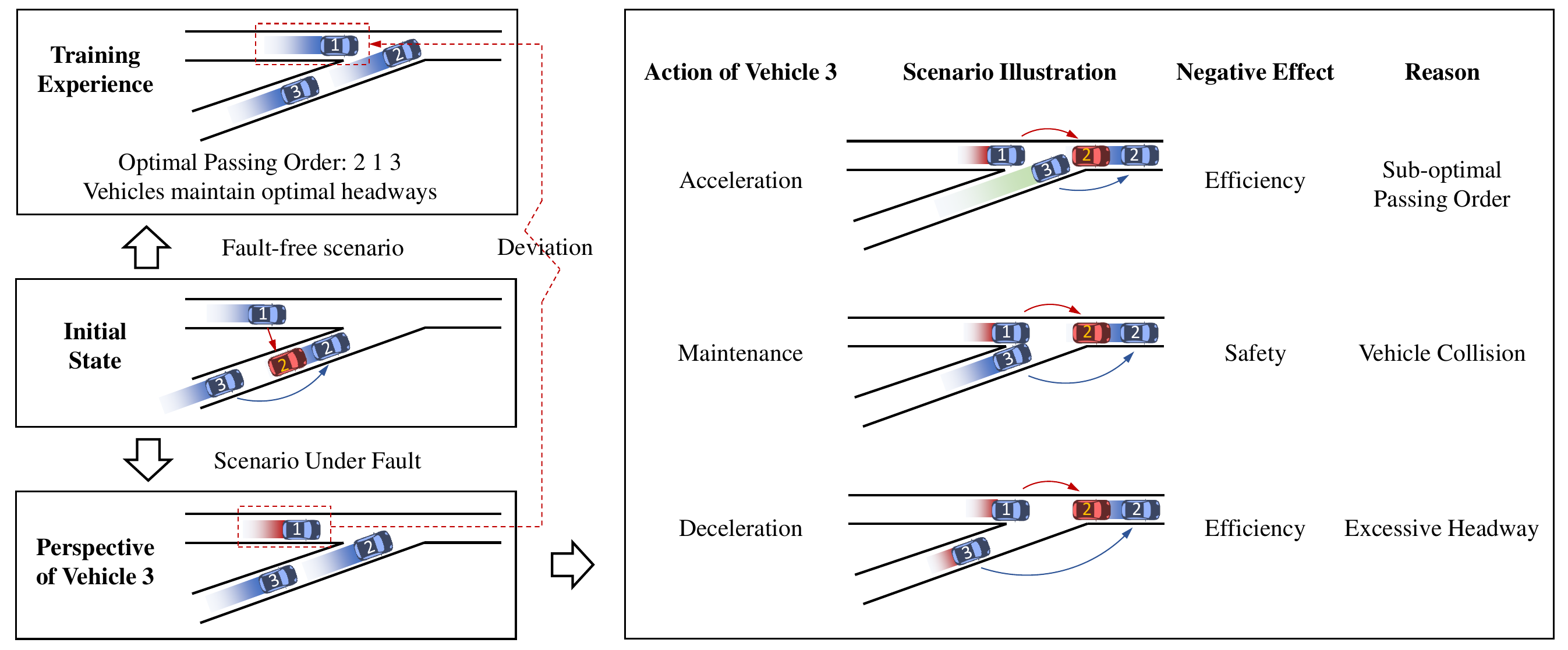}
\caption{Vehicle 1 suddenly misobserves the position of vehicle 2, perceiving it at the location marked in red (Left Middle). In the fault-free scenario, all vehicles would maintain velocities and proceed according to the original cooperative strategy (Left Top). However, influenced by the fault, vehicle 1 chooses to decelerate and yield, which is contrasted with the training experience from the perspective of vehicle 3 (Left Bottom). Further, vehicle 3's action becomes unpredictable due to deviations in vehicle 1's behavior. Regardless of vehicle 3's action, traffic efficiency or safety may be affected for different reasons (Right). The blue, green, and red colors behind vehicles indicate their maintenance, acceleration and deceleration behaviors respectively.}
\label{fig:intro}
\end{figure*}

With the rapid advancement of Reinforcement Learning (RL) technology, its application in the autonomous driving domain has become increasingly prevalent \cite{9351818}. For cooperative decision-making problems, MARL demonstrates significant advantages, exhibiting powerful exploration capabilities and the potential to handle complex environmental scenarios \cite{MARLAV24, lv2025local, app11114948}. However, the high performance of vehicles' policies relies on idealized environmental assumptions. If potential communication or perception faults are overlooked during training, vehicles will likely struggle to respond effectively when encountering unforeseen faults during execution due to insufficient experience and unprepared policies \cite{shi2025ft}. In on-ramp merging scenarios, if a vehicle misobserves the positions and velocities of the surrounding vehicles, it is highly likely to compromise traffic safety or efficiency, especially at the merging point, as illustrated in Fig. \ref{fig:intro}. Regrettably, to the best of our knowledge, current MARL studies for cooperative driving rarely consider the impact mechanisms of potential perception or communication faults. The lack of fault tolerance against such potential disturbances has become a critical bottleneck hindering the practical deployment of MARL-based cooperative driving algorithms.

Considering faults during the training process is necessary for improving fault-tolerant ability. Furthermore, even intentionally incorporating fault simulations presents significant challenges. The first challenge is to generate perturbations that effectively stress the policy during training. Compared to rule-based systems, the impact of observation perturbations on MARL's neural networks is harder to model intuitively \cite{Gleave2020Adversarial}, owing to their black-box and data-driven characteristics. Consequently, random or manual fault injection mostly introduces perturbations with inappropriate magnitudes which have negligible impact, leading the training process to inadequately cover challenging key faults. The second challenge is to equip vehicles with the capability to mitigate the impact of corrupted observations. This stems from the inherent uncertainty under faults: a vehicle cannot determine if its own observations are perturbed, while its perceptions of other vehicles' critical states such as positions and velocities may also be erroneous. This dual uncertainty creates a dilemma: blindly trusting faulty data may lead to unsafe decisions, while excessive distrust forces overly conservative behaviors, significantly compromising traffic efficiency. 

To address these challenges, this paper proposes \emph{\underline{O}bservational \underline{F}ault-\underline{T}olerant \underline{MARL}} (OFT-MARL) for CAVs in on-ramp merging scenarios. OFT-MARL incorporates two key modules, forming a defensive-offensive synergy: the fault injection agent hardens the policy during training while the fault-tolerant vehicle agent shields it, resulting in a fault-tolerant system. OFT-MARL is designed as follows: 

To tackle the first challenge of effective perturbation, we introduce a global fault injection agent. This agent aggregates observations and fault information from all vehicles, and calculates the most disruptive observation perturbation strategy based on global situation analysis. The fault injection agent is co-trained concurrently with the vehicles' policies. By continuously exposing the vehicle agents to these optimally adversarial perturbations, their policy's fault-tolerant capabilities are substantially enhanced through adversarial learning. 

Simultaneously, to address the second challenge brought about by corrupted observations, we design the vehicle agent with an intrinsic fault diagnosis capability. This capability is grounded in the spatio-temporal correlation inherent in vehicle operational data. For instance, vehicle trajectories should exhibit continuity and smoothness, without frequent occurrences of sharp acceleration or deceleration during normal driving. By analyzing the temporal characteristics of state sequences, the vehicle can discriminate data that significantly deviates from these typical dynamics and subsequently reconstruct credible estimates of the true observations. This provides vehicles with credibility assessments and correction references, thereby assisting them in forming more reasonable and efficient policies. 

The rest of the paper is organized as follows. Section \ref{sec:Related Works} reviews related work, followed by background knowledge in Section \ref{sec:Background}. Section \ref{sec:Methods} details the proposed fault-tolerant MARL approach for CAVs. Section \ref{sec:Experiments} showcases and analyzes the experimental results. Finally, Section \ref{sec:Conclusion} concludes the paper and suggests future research directions.

\section{Related Works}
\label{sec:Related Works}

\subsection{Cooperative Driving}
Traditional cooperative driving methods are typically achieved by establishing explicit optimization objectives, designing right-of-way allocation rules, and implementing trajectory planning \cite{8867945, LONG2023411, zhang2025optim, wang2025optim}. However, traditional methods can be limited in scalability and adaptability in highly dynamic and complex scenarios. This has motivated the adoption of MARL, which excels at learning cooperative strategies directly from interaction data.

To tackle various practical challenges in traffic, researchers have developed solutions built upon foundational frameworks. Ramp merging scenario is one of the typical scenario for algorithm verification. For instance, Chen \emph{et al.} \cite{10159552} incorporated a priority-based safety supervisor into MA2C to enhance safety considerations. Pan \emph{et al.} \cite{pan2025trust} proposed Trust-MARL based on MASAC to control mixed traffic flows on ramps. Chen \emph{et al.} \cite{chen2021graph} integrated graph neural networks with MADQN to handle dynamic vehicle numbers and rapidly expanding joint action spaces for cooperative lane change control during merging. In this work, to achieve precise velocity control, we enhance the fault tolerance of the MADDPG \cite{MADDPG} algorithm, which is suitable for continuous action control, and validate it in ramp scenarios.

\subsection{Fault Tolerance for Cooperative Driving}

Classical fault-tolerant methods for cooperative driving are typically implemented through scheduling rule design or game-theoretic optimization. Pei \emph{et al.} \cite{pei2023Ft} constructs a rule-based fault-tolerant cooperative driving strategy. He \emph{et al.} \cite{he2024communication} proposed a global planning and local gaming framework. Liu \emph{et al.} \cite{liu2023fault} introduced a method for real-time decision adjustments, where faulty and normal vehicles can execute distinct decision models in a distributed manner. Ma \emph{et al.} \cite{MA2021100312} transformed robust autonomous intersection control into a weighted maximal clique problem with restrictions and employed a heuristic algorithm to explore the solution space. Such traditional methods possess certain advantages in fault handling, for their relatively fixed and predictable decision rules. To address potential input biases in RL-based vehicle policy networks, He \emph{et al.} utilized Bayesian optimization to approximate input perturbations for black-box vehicle policies \cite{9750867}, while applied gradient descent for white-box scenarios with known policy gradients \cite{9994638}. However, most studies about robust RL-based vehicle policies focus on single-vehicle settings, with limited research on multi-vehicle interactions.

\section{Background}
\label{sec:Background}

\subsection{Markov Games}
A multi-agent system with $N$ agents can be modeled as decentralized partially observable Markov decision processes (Dec-POMDPs) \cite{Markov94, POMDP98}, defined as $<S,A,T,R,Y,O,\gamma>$, where ${S}$ is the set of states, ${A=\{A_1,...,A_N\}}$ is the set of joint actions, ${T}$ is the transition function, ${O=\{O_1,...,O_N\}}$ is the set of observations, and ${Y=\{Y_1,...,Y_N\}}$ is the set of observation functions. 
At each time step $t$, agent ${i}$ receives a partial observation ${o_{i}=Y_i(s):S\rightarrow O_i}$, takes action ${a_{i}}$ according to policy ${\pi_{\theta_i}}$: ${a_{i}=\pi_{\theta_i}(o_{i}):O_i\rightarrow A_i}$, and receives a reward $r_i=R(s,a_1,...,a_N):S\times A_1 \times ... \times A_N \rightarrow \mathbb{R}$. Therefore, the environment at the next time step can be described as ${s'=T(s,a_{1},...,a_{N})}:S\times A_1\times ...\times A_N \rightarrow S$. 
The objective for each agent is to maximize its expected discounted reward ${\mathbb{E}[R_i]=\mathbb{E}[\sum_{t'=t}^{t_0}\gamma^{t'-t}r_{i,t'}]}$, where ${\gamma\in[0,1]}$ is the discount factor, $r_{i,t'}$ is the reward at time step $t'$ and $t_0$ is the end time step of an episode.

\subsection{Deep Deterministic Policy Gradient (DDPG)}
DDPG \cite{DDPG16} helps an agent to learn a deterministic continuous action $a = \mu_{\theta}(s)$, where $\mu$ is the actor network parameterized by $\theta$. The policy gradient is:
\begin{equation}
	{\nabla_{\theta}J(\theta)}=\mathbb{E}_{s \sim U(\mathcal{D})}[\nabla_{\theta}\mu_{\theta}(s)\nabla_{a}Q^{\phi}(s, a)|_{a=\mu_{\theta}(s)}].
\end{equation}
Here, $U(\mathcal{D})$ denotes experiences sampled from the replay buffer $\mathcal{D}$, which stores tuples $(s,s',a,r)$. $Q^{\phi}(s, a)$ is the Q-value function, also called the critic network. Its parameters $\phi$ is updated via:
\begin{equation}
	\begin{split}
		\mathcal{L}(\phi)=\mathbb{E}_{(s,a,r,s') \sim U(\mathcal{D})}[(Q^{{\phi}}(x,a)-y)^2],\\
		y=r+\gamma Q^{\phi'}(s',a')|_{a'=\mu_{\theta'}(s')},
	\end{split}
\end{equation}
where $\mu_{\theta'}$ is the target policy with delayed parameters $\theta'$.

MADDPG \cite{MADDPG} is a variant of DDPG for multi-agent systems, with centralized Q-value function and decentralized policies. Considering $N$ agents with continuous deterministic policies $\mu=\{\mu_{\theta_1},...,\mu_{\theta_N}\}$, the policy gradient for agent $i$ is:
\begin{equation}
	{\nabla_{\theta_i}J(\theta_i)}=\mathbb{E}_{(x,a) \sim U(\mathcal{D})}[\nabla_{\theta_i}\mu_{\theta_i}(o_i)\nabla_{a_i}Q_i^{\phi_i}(x,a)|_{a_i=\mu_{\theta_i}(o_i)}].
\end{equation}
Here, $Q_i^{\phi_i}(x,a)$ is agent $i$'s Q-value function, which accepts all agents' actions $a=[a_1,...,a_N]$ and state information $x$. 
In this paper, $x$ is defined as the concatenation of all agents' observations $[o_1,...,o_N]$.
$\mathcal{D}$ stores tuples $(x,x',a,r_1,...,r_N)$, encapsulating all agents' experiences. $\phi_i$ is updated via:
\begin{equation}
	\begin{split}
		\mathcal{L}(\phi_i)=\mathbb{E}_{(x,a,r,x') \sim U(\mathcal{D})}[(Q_i^{{\phi_i}}(x,a)-y_i)^2],\\
		y_i=r_i+\gamma Q_i^{\phi'_i}(x',a'_1,...,a'_N)|_{a'_j=\mu_{\theta'_j}(o'_j)}.
	\end{split}
\end{equation}
The DDPG and MADDPG frameworks described above form the foundation of our method. We introduce novel fault-tolerant mechanisms considering vehicle characteristics within the established frameworks to address the vulnerability to observational faults.

\begin{figure*}[htbp]
\centering
\includegraphics[width=0.9\linewidth]{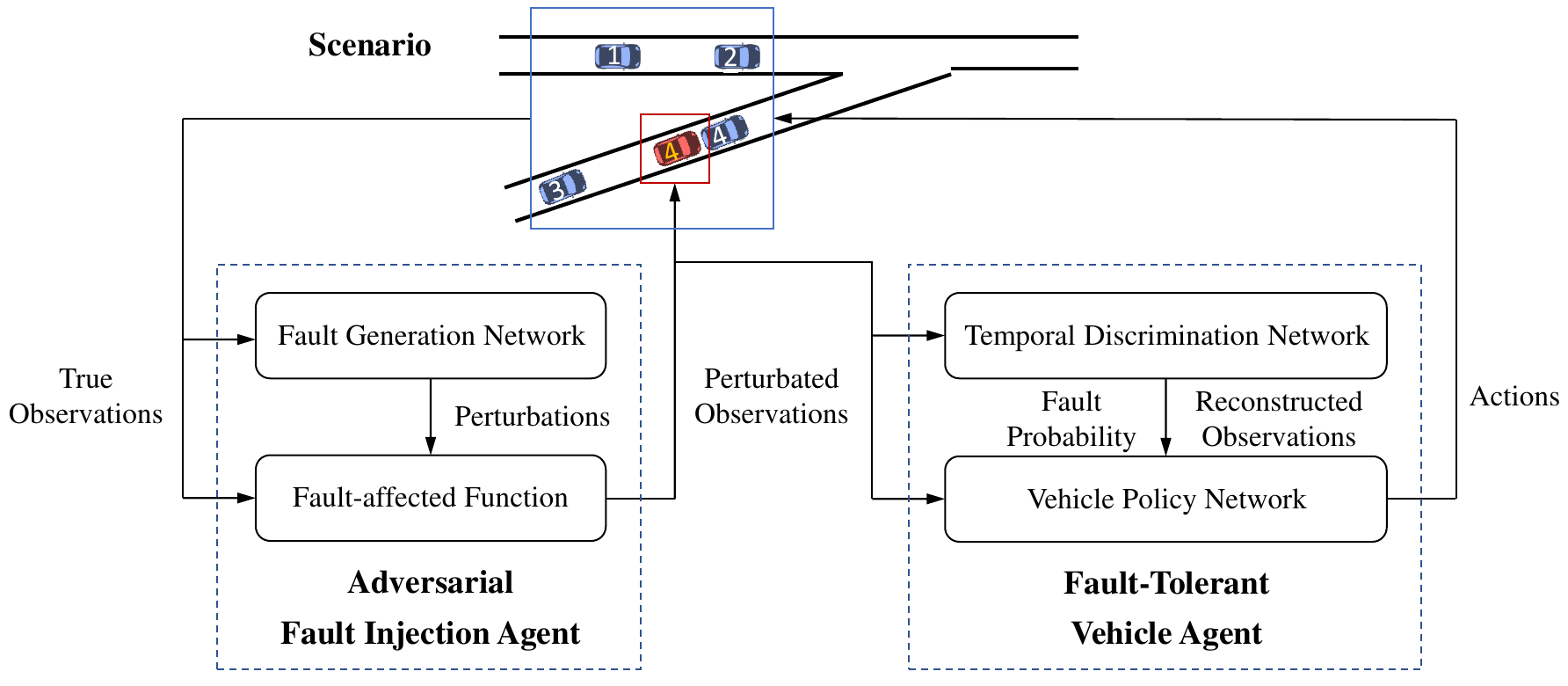}
\caption{Overall framework of OFT-MARL.}
\label{fig:Overall_method}
\end{figure*}

\section{Methods}
\label{sec:Methods}
This section details OFT-MARL designed for the highway on-ramp merging scenario. As illustrated in Fig. \ref{fig:Overall_method}, the framework is composed of the driving scenario along with two interacting agents. The core of our approach lies in the offensive-defensive synergy established between the fault injection agent and the fault-tolerant vehicle agent.

Initially, a problem formulation is established (Section \ref{sec: method0}), followed by the formalization of the observation model and fault model to define the mechanism of observation perturbations (Section \ref{sec: method1}). Subsequently, a fault injection agent is introduced, which adaptively generates strategically disruptive perturbations within bounded constraints from the true observations and reapplies them to the scenario (Section \ref{sec: method2}). The fault-tolerant vehicle agent is equipped with the capability  to leverage sequential dependencies for simultaneous fault detection and observation reconstruction, thereby generating reliable actions (Section \ref{sec: method3}). Finally, all components are integrated via a joint training framework (Section \ref{sec: method4}), with details of each component depicted in Fig. \ref{fig:method}.

\subsection{Problem Formulation}
\label{sec: method0}
The scenario addressed in this study is a highway on-ramp merging configuration. Denote the fleet size by $N$, where each vehicle $i \in \{1,\dots,N\}$ has a state vector $s_i \in \mathbb{R}^d$. Normally, vehicle $i$ gets true observation $o_{i}=Y_i(s)$. However, communication or perception faults impair observation function $Y_i$, affecting the accuracy of the observation and causing $o_i$ to diverge from the ground-truth state $s$. To facilitate a clear representation, we denote \( o_i \) and \( Y_i \) as the pristine observation and observation function, respectively, while \( \hat{o}_i \) and \( \hat{Y}_i \) represent the potentially perturbed observation and observation function. The space of all possible perturbed observations is denoted by \( \hat{O} \). 

To address the problem, our methodology proceeds as follows: first, we formally define \( Y_i \) and \( \hat{Y}_i \) while constraining the domain of \( \hat{O} \); second, we introduce the fault injection method \( f: O \rightarrow \hat{O} \); and finally, we articulate the policy \( \mu: \hat{O}_i \rightarrow A_i \) that maps perturbed observations to actions. These components collectively form the foundational issues examined in the subsequent sections.

\subsection{Observation Modeling under Faults}
\label{sec: method1}
We formally define the observation model under normal and fault conditions, and establish perturbation constraints.

\begin{figure*}[htbp]
\centering
\includegraphics[width=0.95\linewidth]{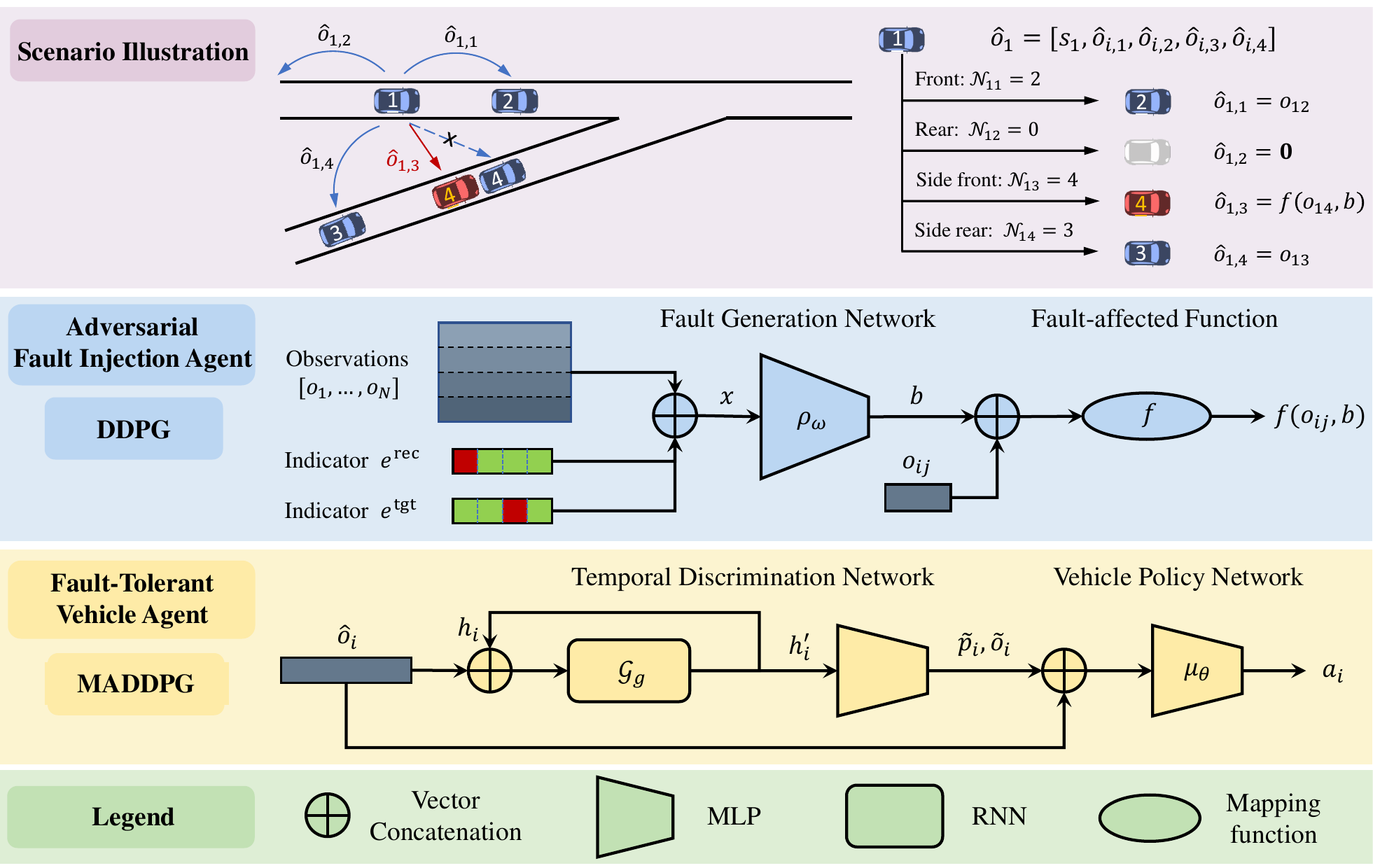}
\caption{Schematic diagram of OFT-MARL. Top: Illustration of a on-ramp merging scenario case (left) and the composition of the observation for vehicle 1 within the scenario (right). Middle: Workflow of the adversarial fault injection agent, which takes the observations of all vehicles and two fault indicators as input, processes them through the fault generation network to output perturbation magnitude $b$, and then transforms the true observation $o_{ij}$ into fault-affected observation $f(o_{ij},b)$ via the fault-affected function. Bottom: Workflow of the fault-tolerant vehicle agent, which takes potentially perturbed observation $\hat{o}_i$ along with the hidden state $h_i$ from the previous timestep (via the GRU), outputs the hidden state $h'_i$ for the next timestep, further processes it through an MLP network to generate fault predictions $\tilde{p}_{i}$ and observation estimates $\tilde{o}_{i}$, and finally combines these with the observation $\hat{o}_i$ to produce acceleration action $a_i$ through the action output network.}
\label{fig:method}
\end{figure*}

Each vehicle's policy depends primarily on neighboring vehicles. To reduce computational and communication overhead, vehicle $i$ observes only its own state and those of $m$ surrounding vehicles \cite{10159552,10367764}. 
Let $\mathcal{N}_i = [ \mathcal{N}_i^1, \dots, \mathcal{N}_i^m ]$ be the neighbor index array of vehicle $i$, where $\mathcal{N}_i^j \in \{ 0, 1, \dots, N \}, j = 1, \dots, m$. The neighbor may not exist due to spatial distribution or insufficient vehicle density, in which case we set $\mathcal{N}_i^j = 0$. As depicted in Fig. \ref{fig:method} Top, $\mathcal{N}_1 = [2,0,4,3]$ shows the distribution of vehicles at 4 relative positions around vehicle 1. The true observation vector $o_i \in \mathbb{R}^{(m+1)d}$ is constructed as:
\begin{equation}
o_i = Y_i(s) = \left[ s_i,  o_{i,1}, \dots, o_{i,m} \right],
\end{equation}
where $s_i$ is the state of vehicle $i$ and $o_{i,j} \in \mathbb{R}^d$ is the observation vector from vehicle $i$ to its $j$-th neighbor:
\begin{equation}
o_{i,j} = 
\begin{cases} 
\mathbf{0}_d, & \mathcal{N}_i^j = 0, \\  
o_{i\mathcal{N}_i^j}, & \mathcal{N}_i^j > 0.
\end{cases}
\end{equation}
Here, $\mathbf{0}_d$ denotes a $d$-dimensional zero vector when neighbor $j$ is absent and $o_{ij}$ is the true observation from vehicle $i$ to vehicle $\mathcal{N}_i^j$.

The observation vector considering potential faults $\hat{o}_i \in \mathbb{R}^{(m+1)d}$ is constructed as:
\begin{equation}
\label{eq:hat_o_1}
\hat{o}_i = \hat{Y}_i(s) = \left[ s_i,  \hat{o}_{i,1}, \dots, \hat{o}_{i,m}, \right],
\end{equation}
where $\hat{o}_{i,j}$ is the observation vector from vehicle $i$ to its $j$-th neighbor, which may contain potential perturbations:

\begin{equation}
\label{eq:hat_o_2}
\hat{o}_{i,j} = 
\begin{cases} 
\mathbf{0}_d, & \mathcal{N}_i^j = 0, \\ 
f(o_{i\mathcal{N}_i^j}, b), & \mathcal{N}_i^j > 0 \text{ under fault condition}, \\ 
o_{i\mathcal{N}_i^j}, & \mathcal{N}_i^j > 0 \text{ under normal condition}.
\end{cases}
\end{equation}
Here, $f(o_{i\mathcal{N}_i^j}, b)$ presents the fault-affected observation. $b$ is the deviation magnitude, applied to the true observation $o_{i\mathcal{N}_i^j}$ through fault-affected function $f$. The function will be detailedly described in Section \ref{sec: method2}.

To bound fault magnitudes, we define an $l$-norm ball $B(o)$ \cite{zhang2020robust, he2023robust}. The perturbed observation $f(o, b)$ should lie within the ball centered at $o$:
\begin{equation}
B(o) := \left\{ f(o, b) \in \mathbb{R}^d : \| f(o, b) - o \|_l \leq \epsilon \right\},
\end{equation}
where $\epsilon > 0$ is the perturbation budget.
This constraint ensures the problem remains well-posed, as excessive perturbations provide limited practical utility for policy decisions and are readily detectable in real-world scenarios.

\subsection{Adversarial Fault Injection Agent}
\label{sec: method2}
Although fault mechanisms are defined, determining optimal perturbation amplitudes across state dimensions remains challenging. Indiscriminate random fault injection may cause vehicles to frequently learn responses to minor faults while remaining vulnerable to critical faults. Meanwhile, it is difficult to manually set up valid faults, for the impact of observation deviations on RL-based policies is challenging to analyze theoretically. To address the problem, we introduce a global fault injection agent that adaptively introduces disruptive faults during training.

We impose a constraint on the fault injection agent: it cannot arbitrarily choose which vehicle receives a faulty observation or which neighbor's state is perturbed. Instead, these are randomly selected. As exemplified in Fig. \ref{fig:method} (Top), it is randomly determined that the fault recipient vehicle 1 misobserves the fault target vehicle 4. After selecting fault recipient and target vehicles, indicators presenting the fault information and observations of all vehicles together contain the requisite information for finding the effective perturbations from a system-wide global perspective. 

As illustrated in Fig. \ref{fig:method} (Middle), the workflow of the fault injection agent involves three key steps: \emph{i)} aggregating global observations and fault indicators to form the input vector $x$; \emph{ii)} processing $x$ through the policy network to compute perturbation magnitude $b$; and \emph{iii)} using a fault-affected function to apply $b$ to the true observation. 

Formally, the input vector $x = [o_1, \dots, o_N, e^{\text{rec}}, e^{\text{tgt}}]$ concatenates: \emph{i)} All vehicle observations $o_i$; \emph{ii)} Fault recipient indicator $e^{\text{rec}}$; \emph{iii)} Fault target indicator $e^{\text{tgt}}$. Here, $e^{\text{rec}}$ indicates vehicle $i$ which gets perturbed observations and $e^{\text{tgt}}$ indicates vehicle $\mathcal{N}_i^j$ which vehicle $i$ misobserves. Both are one-hot vectors:
\begin{equation}
    \begin{split}
    e^{\text{rec}} = [\delta_{ki}]_{k=1}^{N} = \begin{cases} 
    1, & k = i \\
    0, & \text{otherwise}
    \end{cases},\\
    e^{\text{tgt}} = [\delta_{kj}]_{k=1}^{m} = \begin{cases} 
    1, & k = j \\
    0, & \text{otherwise}
    \end{cases},
    \end{split}
\end{equation}
where $\delta_{ki}$, $\delta_{kj}$ are Kronecker delta functions.

When vehicle $i$ misobserves neighbor $\mathcal{N}_i^j$, the input-output relationship is:
\begin{equation}
b = \rho_{\omega}(x),
\end{equation}
where $b \in \mathbb{R}^{d_0}$ with $d_0 \leq d$ denotes perturbations applied to $d_0$ state dimensions of $o_{i\mathcal{N}_i^j}$. Here $\rho$ represents the fault injection agent's policy network with parameters $\omega$.

In this paper, we adapt a basic add function as the fault-affected function. The $B(o)$ ball is a $\infty$-norm ball. Thus, 
\begin{equation}
    \label{eq:add_func}
    f(o_{ij}, b) = o_{ij} + \mathrm{clip} \left( b, -\epsilon, \epsilon \right).
\end{equation}
Here, $\mathrm{clip} \left( b, -\epsilon, \epsilon \right)$ means each dimension of $b$ is restricted in $[-\epsilon, \epsilon]$.

\subsection{Fault-Tolerant Vehicle Agent}
\label{sec: method3}
Observation perturbations pose a significant threat to vehicle policies, particularly those based on conventional Multi-Layer Perceptron (MLP) architectures. This vulnerability stems from the fundamental memoryless nature of MLPs as feedforward networks. While MLPs are well-suited for idealized fully observable Markovian environments where decisions can be made optimally based on accurate current observations, they become critically inadequate in the presence of faults. Specifically, they lack the capability to detect observational faults or extract diagnostic information from corrupted observations.

This inherent limitation severely compromises the safety and operational efficiency of vehicle policies when observation perturbations occur. Fortunately, vehicular operational data exhibit strong spatio-temporal correlations. For instance, vehicle kinematics are bound by physical constraints, trajectories demonstrate continuity and smoothness, and driving behaviors generally adhere to established traffic norms. These consistent temporal patterns provide a viable means to mitigate perturbations by enabling the discrimination of potential faults and the reconstruction of credible observation estimates.

We therefore introduce a temporal discrimination network to equip the fault-tolerant agent with these capabilities. This network is implemented using a Gated Recurrent Unit (GRU), an advanced version of RNNs that utilizes sophisticated gating mechanisms to regulate information flow over extended sequences. The GRU's hidden state, which contains historical information, precisely suits the temporal characteristics of vehicle data. When abrupt deviations manifest in observation streams, this network identifies compromised observation dimensions and generates relatively reliable estimates. Augmenting observation vectors with these outputs significantly enhances fault tolerance.

As illustrated in Fig.~\ref{fig:method} (Bottom), the workflow of the proposed fault-tolerant vehicle agent involves three steps: \emph{i)} updating the hidden state $h_i'$ encoding historical information; \emph{ii)} generating fault probability $\tilde{p}_i$ and the reconstructed observation estimate $\tilde{o}_i$ from $h_i'$; \emph{iii)} outputting the vehicle action $a_i$ through the policy network.

For vehicle $i$, the temporal discrimination network $\mathcal{G}$ with parameters $g$ operates as:
\begin{equation}
\label{eq:gru}
\begin{gathered}
    \tilde{p}_{i}, \tilde{o}_{i} = \mathcal{G}_g (\hat{o}_{i}, h_{i}), \\
    \begin{cases}
        h'_{i} = \sigma(\hat{o}_{i}, h_{i}), \\
        \tilde{p}_{i} = \text{sigmoid}(W_p h'_{i} + b_p), \\
        \tilde{o}_{i} = W_o h'_{i} + b_o.
    \end{cases}
\end{gathered}
\end{equation}
where $h'_{i}$ and $h_{i}$ denotes the hidden states for the next and the current timestep, with $\tilde{p}_{i} \in [0,1]^m$ indicating fault probability of each neighbor, and $\tilde{o}_{i} \in \mathbb{R}^{d_0}$ representing the reconstructed observation. $\sigma$ denotes the GRU and $\text{sigmoid}$ denotes the activation function. $W_p, W_o$ and $b_p, b_o$ represents the weights and biases of the linear layer respectively.

With the help of fault identification and observation prediction, the policy network processes the augmented inputs to output the acceleration action $a_{i}$:
\begin{equation}
a_{i} = \mu_\theta\left( \hat{o}_{i}, \tilde{p}_{i}, \tilde{o}_{i} \right),
\end{equation}
where $\mu$ is the policy network with parameters $\theta$ shared across all vehicles.

\begin{algorithm*}[h]
    \caption{OFT-MARL}    
    \label{alg:training_framework}
    \begin{algorithmic}[1]
    \REQUIRE Batch size $k$, replay buffers $\mathcal{D}_\text{v}$ (vehicle agent) and $\mathcal{D}_\text{f}$ (fault injection agent), number of vehicles $N$, maximum training episodes $E$, maximum steps per episode $S$, network update frequency $K$
    
    \STATE Initialize step counter $\text{Total\_steps} \gets 0$. Randomly initialize the vehicle agent critic network $Q_i^{\psi_i}$, the vehicle agent actor network $\mu_{\theta}$, the fault injection critic network $Q^{\phi}$, the fault injection actor network $\rho_{\omega}$, and the temporal discrimination network $\mathcal{G}_g$.
    
    \FOR {episode $e = 1$ to $E$}
        \STATE Initialize environment state, fault configuration, and GRU hidden states for all vehicles
        \FOR {step $t = 1$ to $S$}
            \IF {fault condition is triggered}
                \STATE Compute perturbation: $b \gets \rho_{\omega}(x)$
            \ELSE
                \STATE $b \gets 0$ (no perturbation)
            \ENDIF
            
            \STATE Generate perturbed observations $\hat{o}_i$ using Equ. (\ref{eq:hat_o_1}), (\ref{eq:hat_o_2}) and (\ref{eq:add_func})
            \STATE Compute temporal features: $(\tilde{p}_{i}, \tilde{o}_{i}, h'_{i}) \gets \mathcal{G}_g(\hat{o}_{i}, h_{i})$ via Equ. (\ref{eq:gru})
            \STATE Select actions: $a_{i} \gets \mu_\theta(\hat{o}_{i}, \tilde{p}_{i}, \tilde{o}_{i})$ for each vehicle $i$
            \STATE Execute actions, observe next state $s'$ and rewards $r_i$
            \STATE Store transition $(x,b,r',x')$ in $\mathcal{D}_\text{f}$ if fault has occurred
            \STATE Store vehicle experiences $(o,\hat{o},a,r,o',h,h')$ in $\mathcal{D}_\text{v}$
            \STATE $\text{Total\_steps} \gets \text{Total\_steps} + 1$
            
            \IF {$\text{Total\_steps} \mod K = 0$} 
                \FOR {vehicle $i = 1$ to $N$}
                    \STATE Sample mini-batch of $k$ transitions from $\mathcal{D}_\text{v}$
                    \STATE Update temporal network $\mathcal{G}_g$ via Equ. (\ref{eq:tmp_net})
                    \STATE Update critic $Q_i^{\psi_i}$ via Equ. (\ref{eq:veh_critic})
                    \STATE Update actor $\mu_\theta$ via Equ. (\ref{eq:veh_actor})
                \ENDFOR
                
                \STATE Sample mini-batch of $k$ transitions from $\mathcal{D}_\text{f}$
                \STATE Update fault injection critic $Q^{\phi}$ via Equ. (\ref{eq:adv_critic})
                \STATE Update fault injection actor $\rho_{\omega}$ via Equ. (\ref{eq:adv_actor})
                \STATE Update target networks parameters $\psi'_i$, $\theta'$, $\phi'$, $\omega'$
            \ENDIF
            
            \STATE Advance to next state: $o \gets o'$, $h \gets h'$
        \ENDFOR
    \ENDFOR
    \end{algorithmic}
\end{algorithm*}

\subsection{Joint Training Framework}
\label{sec: method4}
The joint training framework is characterized by a co-evolutionary dynamic between the two agents. As illustrated in Fig. \ref{fig:Overall_method}, the fault injection agent's perturbations directly affect the vehicle observations, which in turn influence both the temporal network's estimations and the policy network's decisions. Conversely, the vehicle policy's resilience against perturbations shapes the fault injection agent's learning objective. This co-evolutionary process creates an offensive-defensive synergy where each agent continuously adapts to the other's strategies, ultimately leading to fault-tolerant policy learning.

This synergy of the two agents is implemented through three interconnected network components that undergo joint optimization using distinct learning paradigms, concluded in Table \ref{tab:train}. Within the vehicle agent, the temporal discrimination network $\mathcal{G}_g$ is trained via supervised learning, while the actor-critic architecture, comprising the policy network $\mu_\theta$ (actor) and critic networks $\{Q_i^{\psi_i}\}_{i=1}^N$, is optimized using the MADDPG framework. The fault injection agent employs its own actor-critic scheme, where the actor network $\rho_\omega$ generates adversarial perturbations and the critic network $Q^{\phi}$ evaluates their disruptive impact, both trained with DDPG. 

\begin{table}[htbp]
\centering
\caption{Learning paradigms}
\label{tab:train}
\begin{tabular}{lll}
\toprule
\textbf{Networks} & \textbf{Paradigm} & \textbf{Reason} \\
\midrule
$\mathcal{G}_g$ & Supervised learning & explicit supervision signals \\
$Q_i^{\psi_i}, \mu_\theta$ & MADDPG & multi-agent cooperative policy\\
$Q^{\phi}, \rho_\omega$ & DDPG & single-agent adversarial policy\\
\bottomrule
\end{tabular}
\end{table}

The temporal network $\mathcal{G}_g$ is co-trained with reinforcement learning using supervised signals. We minimize the composite mean squared error (MSE) between its outputs and ground-truth references:
\begin{equation}
    \label{eq:tmp_net}
\mathcal{L}(g) = \mathbb{E}_{(o,\hat{o},h) \sim \mathcal{D}_{\text{v}}} \left[ \| \tilde{p}_i - p_i \|_2^2 + \sum_{j=1}^{m} \| \tilde{o}_{ij} - o_{ij} \|_2^2 \right],
\end{equation}
where $p_i \in \{0,1\}^m$ denotes the ground-truth fault indicator vector.

Each vehicle possesses a dedicated critic $Q_i^{\psi_i}$ that receives ground-truth global states $o = (o_{1}, \dots, o_{N})$ to ensure stability:
\begin{equation}
    \label{eq:veh_critic}
    \begin{split}
    \mathcal{L}(\psi_i) & = \mathbb{E}_{(o,a,r,o',h') \sim \mathcal{D}_{\text{v}}} \left[ \left( Q_i^{\psi_i}(o,a) - y_i \right)^2 \right], \\
    y_i & = r_i + \gamma Q_i^{\psi_i'}\big( o', a'_1, \dots, a'_N \big), \\
    a'_j & = \mu_{\theta'}(\hat{o}'_j, \tilde{p}'_{i}, \tilde{o}'_{i}),
    \end{split}
\end{equation}
where $\psi_i'$ and $\theta'$ denote target network parameters. It is worth noting that the fault injection agent $\rho_\omega$ is incorporated in order to compute perturbed next observation $\hat{o}'_j$. The policy gradient update follows:
\begin{equation}
\label{eq:veh_actor}
\begin{split}
\nabla_{\theta} J(\theta) = \mathbb{E}_{(o,\hat{o},a,h) \sim \mathcal{D}_{\text{v}}} \left[\nabla_{\theta} \mu_\theta(\hat{o}_{i}, \tilde{p}_{i}, \tilde{o}_{i}) \nabla_{a_i} Q_i^{\psi_i}(o,a) \right], \\
a_i=\mu_{\theta}(\hat{o}_{i}, \tilde{p}_{i}, \tilde{o}_{i}).
\end{split}
\end{equation}

The fault injection agent's objective is to maximize policy disruption by minimizing collective rewards:
\begin{equation}
r' = -\sum\nolimits_{i=1}^{N} r_i.
\end{equation}

The fault injection agent possesses a critic network $Q$ with parameters $\phi$, updated to minimize the temporal difference loss:
\begin{equation}
    \label{eq:adv_critic}
    \begin{split}
        \mathcal{L}(\phi) = \mathbb{E}_{(x,b,r',x') \sim \mathcal{D}_{\text{f}}} \left[ \left( Q^{\phi}(x,b) - y \right)^2 \right], \\
        y = r'+\gamma Q^{\phi'}(x',b')|_{b'=\rho_{\omega'}(x')}.
    \end{split}
\end{equation}
where $\phi'$ and $\omega'$ denote target network parameters. The policy gradient update is:
\begin{equation}
    \label{eq:adv_actor}
    \nabla_{\omega} J(\omega) = \mathbb{E}_{x \sim \mathcal{D}_{\text{f}}} \left[ \nabla_{\omega} \rho_{\omega}(x) \nabla_{b} Q^{\phi}(x,b) \big|_{b=\rho_{\omega}(x)} \right].
\end{equation}

The complete algorithm is described in Algorithm \ref{alg:training_framework}.

\section{Experiments}
\label{sec:Experiments}
\subsection{Setup}
\subsubsection{Experiments Overview}

This section conducts experiments to validate the effectiveness of the proposed methodology. We evaluate the contribution of each module by comparing the performance across different experimental configurations, as summarized in \cref{tab:exp_config}.

\begin{table}[htbp]
\centering
\caption{Experimental configurations in two dimensions}
\label{tab:exp_config}
\begin{tabular}{cll}
\toprule
\textbf{Dimension} & \textbf{Configurations} & \textbf{Description} \\
\midrule
\multirow{3}{*}{\makecell{Fault \\ Injection}} 
  & Fault-free & $b=0$ \\
  & Random fault & $b\sim U(-\epsilon,\epsilon)$ \\  
  & Adversarial fault & $b = \rho_{\omega}(x)$ \\
\midrule
\multirow{3}{*}{\makecell{Policy \\ Training}} 
  & Vanilla MADDPG & Baseline without fault tolerance \\
  & OFT-MARL w/o GRU & Proposed method without GRU \\
  & OFT-MARL & Full proposed method \\
\bottomrule
\end{tabular}
\end{table}

In Subsection~\ref{sec: exp1}, we justify the necessity of accounting for potential faults. Policies are trained using vanilla MADDPG in fault-free scenarios, and are then evaluated under both fault-free and random fault conditions, demonstrating the importance of enhancing fault tolerance.

In Subsection~\ref{sec: exp2}, we introduce random faults during training for all three policy methods. Subsequently, a fault injection agent is trained against each fixed pre-trained policy. By comparing performance under random faults versus faults from the fault injection agent during testing, we validate the specific role of the proposed fault injection agent.

In Subsection~\ref{sec: exp3}, all three policy methods are trained with the proposed fault injection agent generating adversarial faults. Policies are evaluated to verify the effectiveness of the overall training architecture and the fault-tolerant vehicle agent. Furthermore, policies are tested under diverse fault patterns to assess the generalization capability.

In Subsection~\ref{sec: exp4}, we perform an isolated analysis of the outputs from the temporal network to further confirm its contribution.

\subsubsection{Scenario}
We design a scenario modified from the Highway Environment \cite{highway-env}. As depicted in Fig. \ref{fig:method} Top, the highway merges consist of single-lane main road and on-ramp with $N=4$ vehicles. Each vehicle $i$ observes $m=4$ neighboring vehicles:
\begin{align*}
&\mathcal{N}_i^1: \text{front vehicle} \\
&\mathcal{N}_i^2: \text{rear vehicle} \\
&\mathcal{N}_i^3: \text{side front vehicle} \\
&\mathcal{N}_i^4: \text{side rear vehicle}
\end{align*}
Thus vehicle $i$'s observation vector $o_i \in \mathbb{R}^{(4+1)\times 4}$ concatenates its own state and relative observations of these neighbors:
\begin{equation}
o_i = [\underbrace{s_i}_{\text{ego}}, \underbrace{o_{i\mathcal{N}_i^1}}_{\text{front}}, \underbrace{o_{i\mathcal{N}_i^2}}_{\text{rear}}, \underbrace{o_{i\mathcal{N}_i^3}}_{\text{side front}}, \underbrace{o_{i\mathcal{N}_i^4}}_{\text{side rear}}].
\end{equation}

State dimensions $d=4$ comprise: (a) existence Boolean, (b) longitudinal position, (c) velocity, and (d) lane ID. Neighbor observations use relative values for position, velocity and lane. Since the existence and lane features are considered unperturbable or difficult to perturb, only the relative position and velocity can be perturbed, which means $d_0=2$.

Vehicles collaborate to achieve safe and efficient merging, with success defined as all vehicles passing a designated point. The reward function combines:
\begin{equation}
r = -1 + r_{\text{goal}} + r_{\text{vel}} + r_{\text{col}},
\end{equation}
where $-1$ prevents vehicles from learning overly conservative local optimal policies, $r_{\text{goal}}$ is a time-discounted task completion bonus, $r_{\text{vel}}$ is the reward for maintaining appropriate velocity, and $r_{\text{col}}$ is the penalty for vehicle collision.

In order to better present the adverse effects of the faults, we increase the randomness of the initial velocities, set strict vehicle headway, and endow vehicles with heterogeneous acceleration characteristics. This prevents trivialization of decision-making tasks and ensures the necessity of the fault tolerance of the vehicle policy.

\subsection{Necessity of Fault tolerance}
\label{sec: exp1}
\begin{figure}[htbp]
    \begin{subfigure}{\linewidth}
        \centering
        \includegraphics[width=0.6\linewidth]{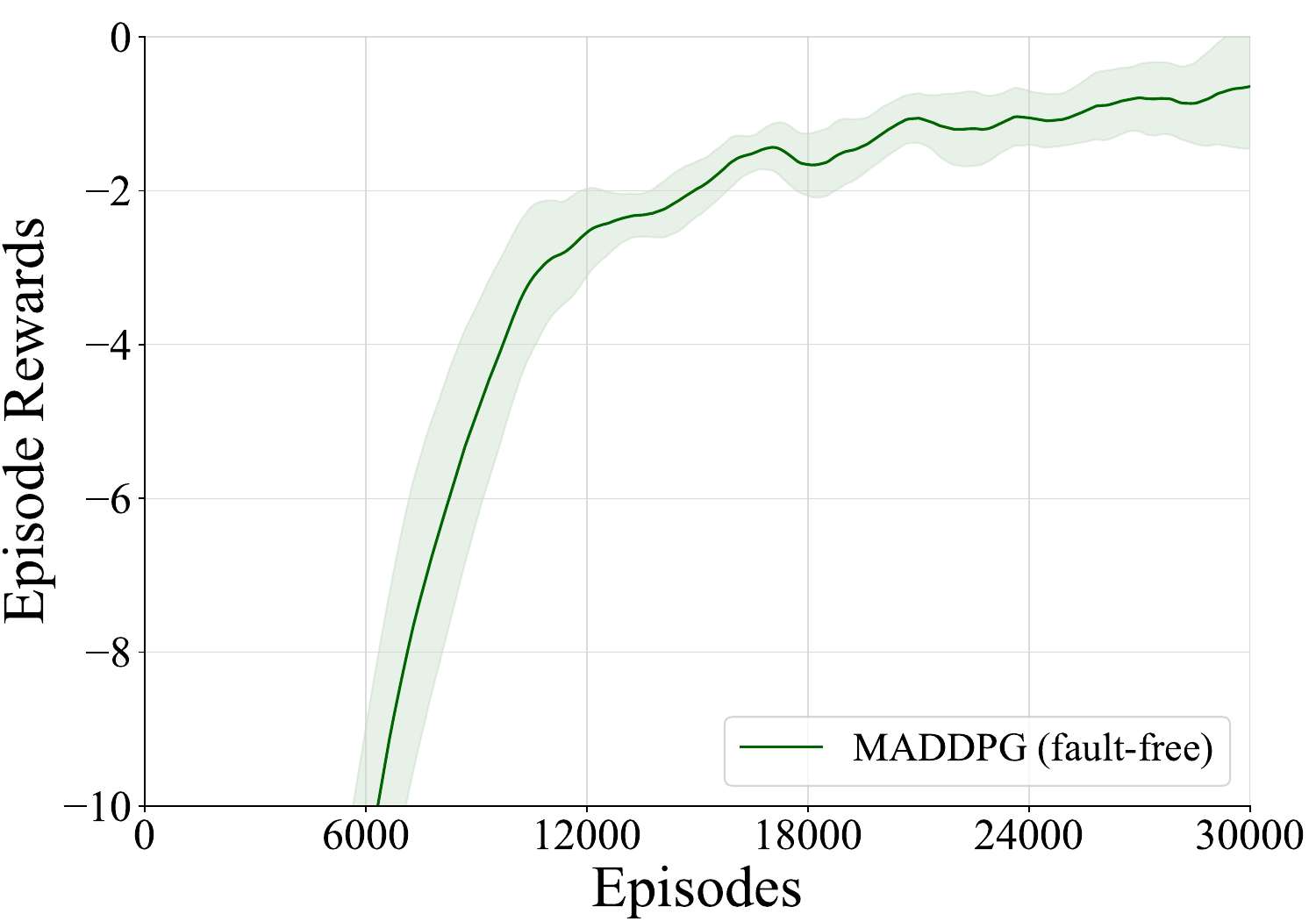}
        \vskip -0.05in
        \caption{}
    \end{subfigure}
    \vspace{0.2cm}
    \begin{subfigure}{\linewidth}
        \centering
        \includegraphics[width=\linewidth]{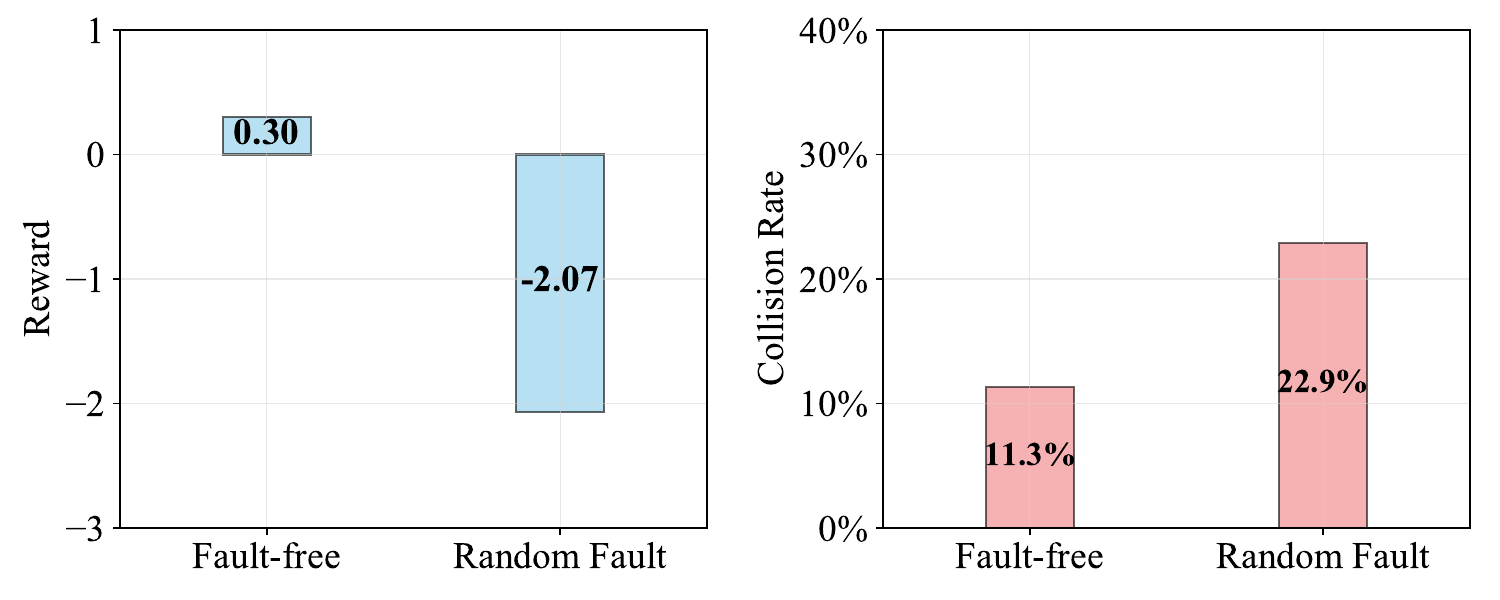}
        \vskip -0.05in
        \caption{}
    \end{subfigure}

\caption{(a) Average reward curves for MADDPG in fault-free scenarios. The error bar is a 95\% confidence interval across 6 runs. (b) Performance decline of methods trained in fault-free scenarios and tested under fault-free and random fault conditions.}
\label{fig:nec_curve}
\end{figure}

We first train vehicle policies using vanilla MADDPG framework without considering potential faults. The training performance is illustrated in Fig. \ref{fig:nec_curve} (a). During evaluation, we test policies trained with 6 different random seeds under two conditions: without faults and with random faults. Results in Table \ref{fig:nec_curve} (b) demonstrate significant performance degradation when random faults are introduced.

\begin{figure}[htbp]
\centering
    \begin{subfigure}{\linewidth}
        \centering
        \includegraphics[width=\linewidth]{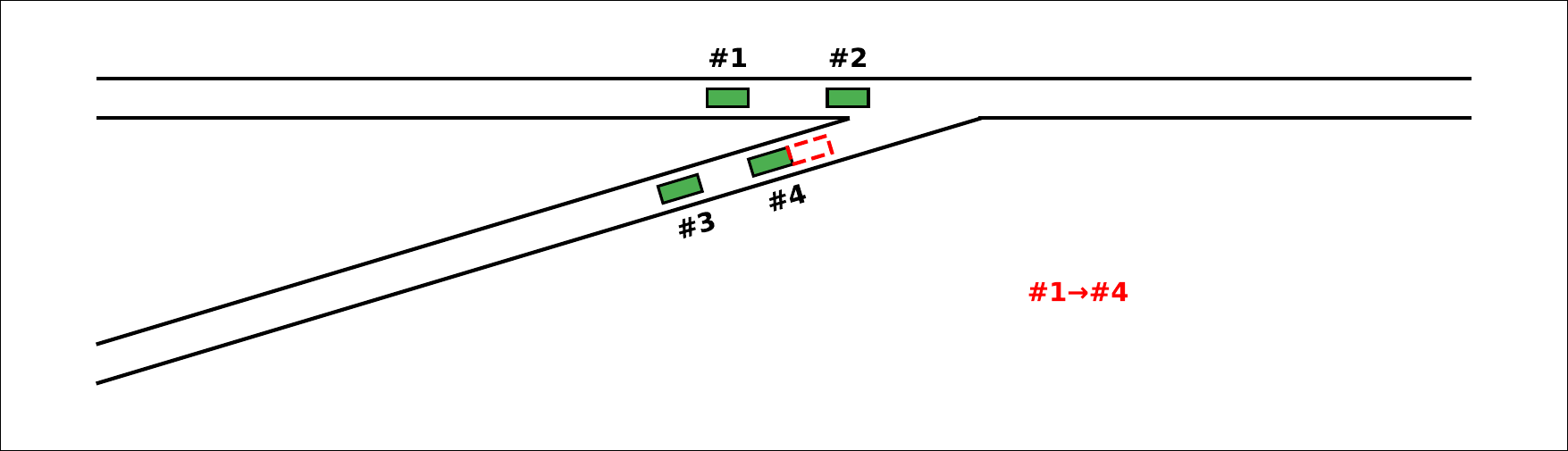}
        \caption{Timestep 9}
    \end{subfigure}
    \begin{subfigure}{\linewidth}
        \centering
        \includegraphics[width=\linewidth]{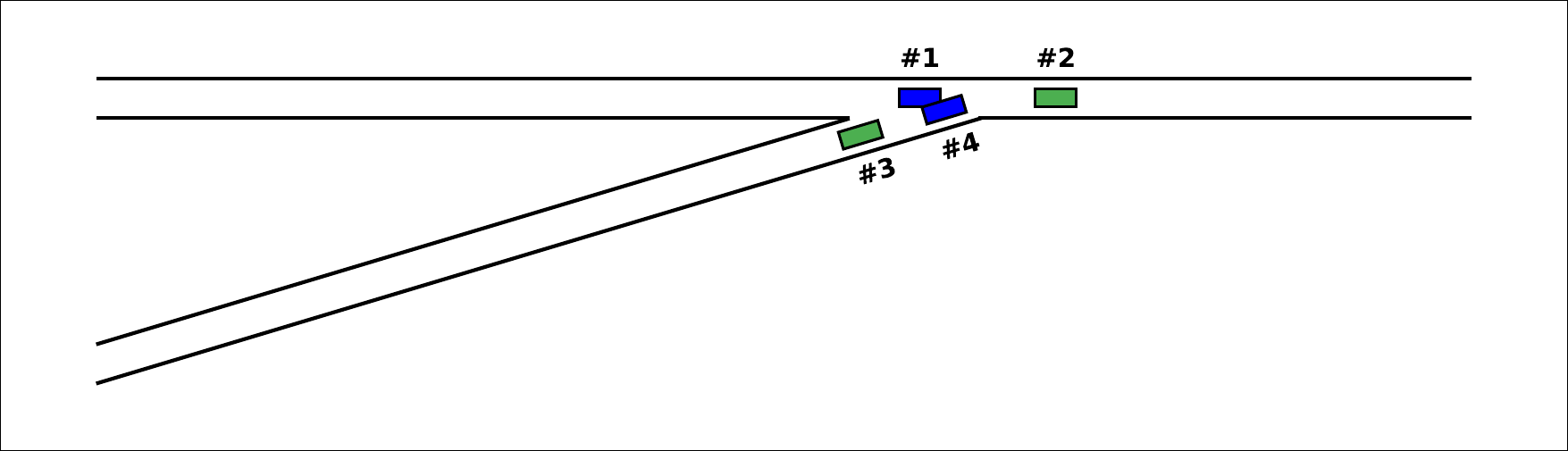}
        \caption{Timestep 11}
    \end{subfigure}

\caption{Influence of random faults.}
\label{fig:nec}
\end{figure}

As exemplified in the test scenario of Fig. \ref{fig:nec}, the true positions of the vehicles are depicted in green. After introducing random faults, at timestep 9 (Fig. \ref{fig:nec} (a)), vehicle 1 receives corrupted observation regarding vehicle 4, falsely perceiving its position at the red dashed bounding box. This perception error leads to failure in recognizing the imminent collision risk and consequently no deceleration is conducted. The collision ultimately occurs at timestep 11 (Fig. \ref{fig:nec} (b)), indicated by the blue highlight.

This experiment substantiates that training processes must account for potential observational deviations, as even random perturbations can critically compromise performance.

\subsection{Effect of the Fault Injection Agent}
\label{sec: exp2}

\begin{figure}[htbp]
\includegraphics[width=\linewidth]{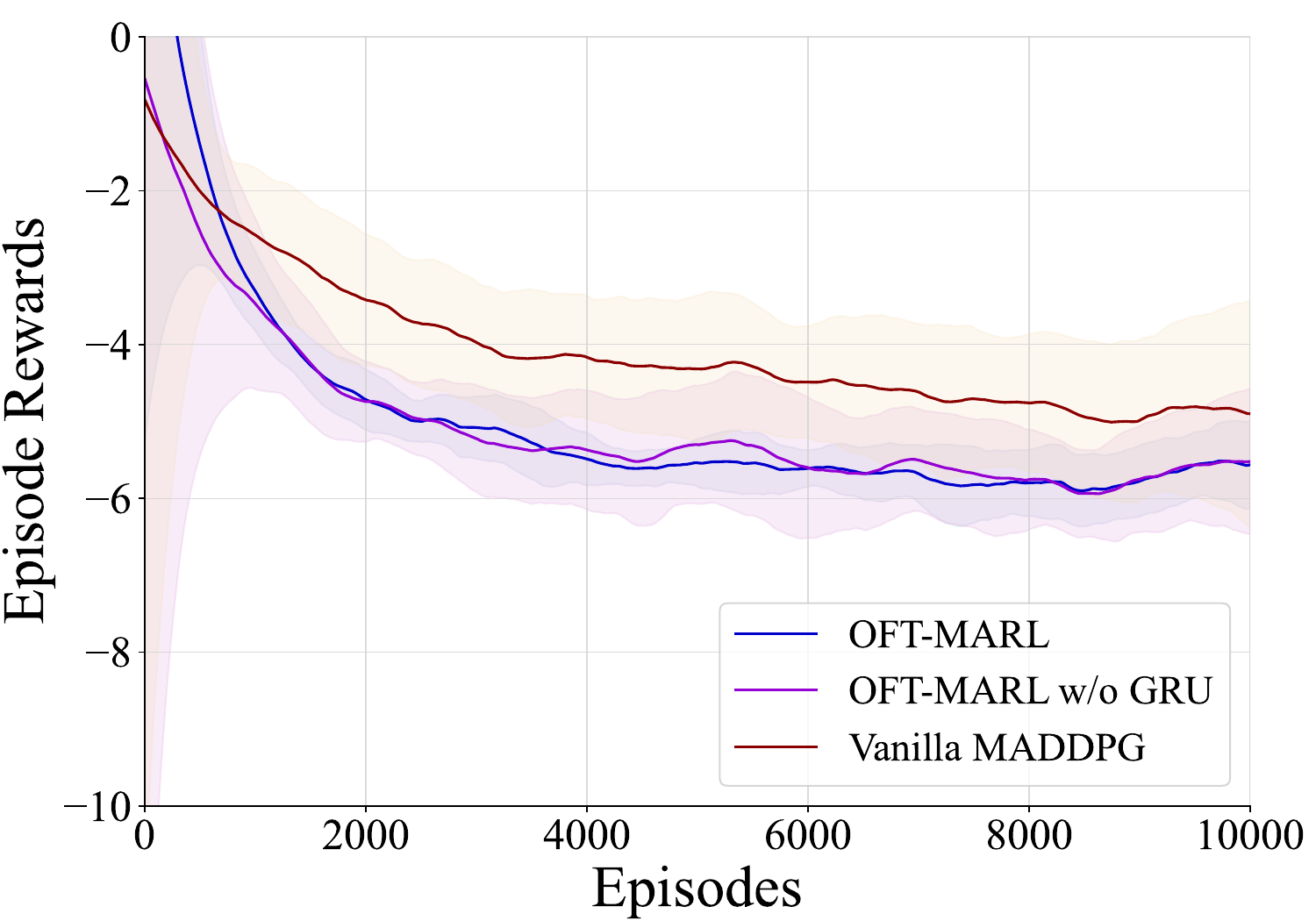}
\caption{Average reward curves for training the fault injection agent with different fixed vehicle policies. The decline in vehicle rewards demonstrates the increasing effectiveness of the adversarial agent. Error bars indicate 95\% confidence interval across 6 runs. }
\label{fig:inj_curve}
\end{figure}

To validate the efficacy of the proposed fault injection agent, it is necessary to demonstrate that introducing merely random faults during training is insufficient. Hence, we first train the policy with random faults and keep it fixed. Then, we introduce the fault injection agent and proceed to train this agent. As illustrated in Fig.~\ref{fig:inj_curve}, the progressively decreasing reward curve during training indicates that the negative impact of faults on the vehicle policy is gradually increasing. The testing results, summarized in Table \ref{tab:inj}, show that while the collision rate changes marginally, the task completion timesteps increase significantly. This phenomenon occurs because the reduction in rewards can be attributed to two primary factors: causing vehicle collisions or impairing traffic efficiency. When the fault injection agent targets traffic efficiency delays, the safety metric may conversely improve. Consequently, the overall collision rate remains relatively unchanged.

\begin{figure}[htbp]
\centering
\begin{subfigure}{\linewidth}
  \centering
  \includegraphics[width=\linewidth]{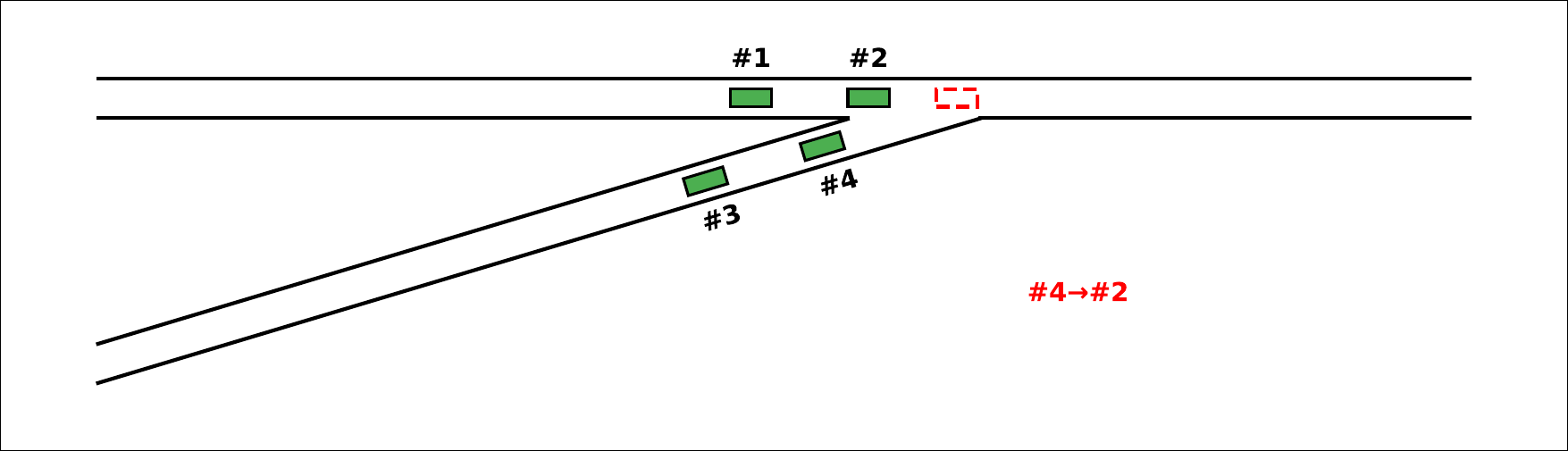}
  \caption{Case 1: timestep 9}
\end{subfigure}
\begin{subfigure}{\linewidth}
  \centering
  \includegraphics[width=\linewidth]{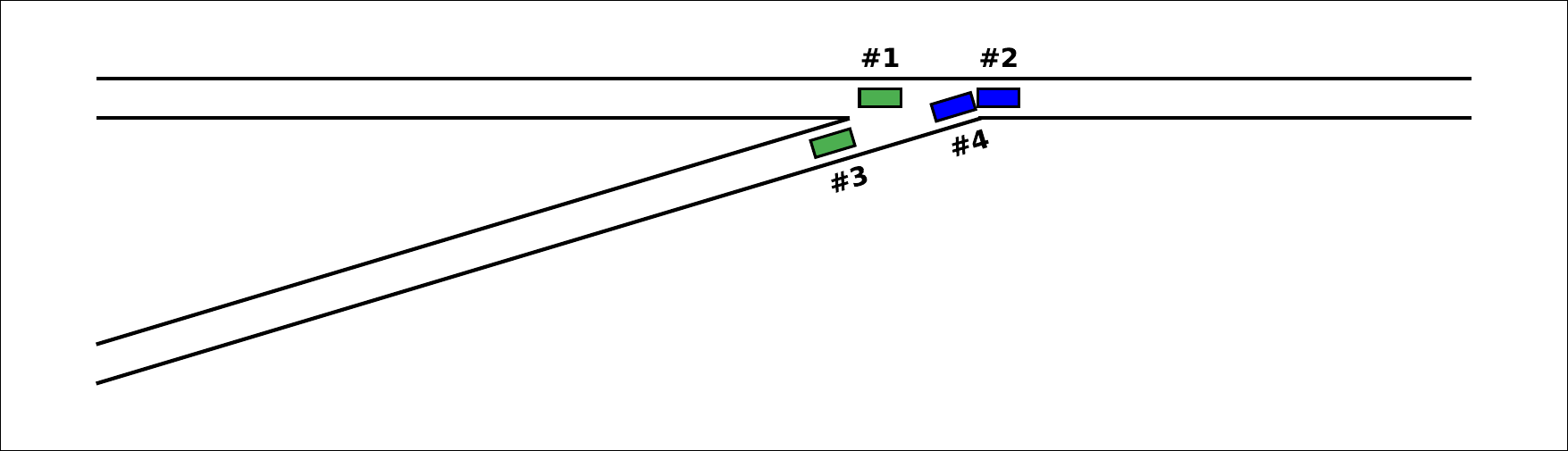}
  \caption{Case 1: timestep 11}
\end{subfigure}
\begin{subfigure}{\linewidth}
  \centering
  \includegraphics[width=\linewidth]{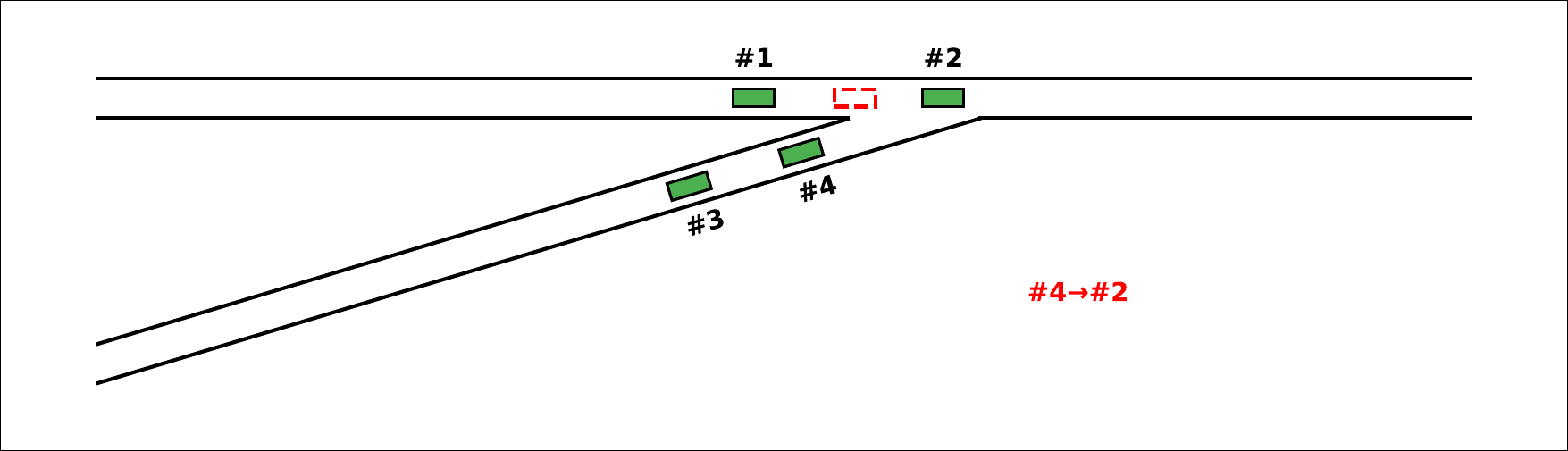}
  \caption{Case 2: timestep 9}
\end{subfigure}
\begin{subfigure}{\linewidth}
  \centering
  \includegraphics[width=\linewidth]{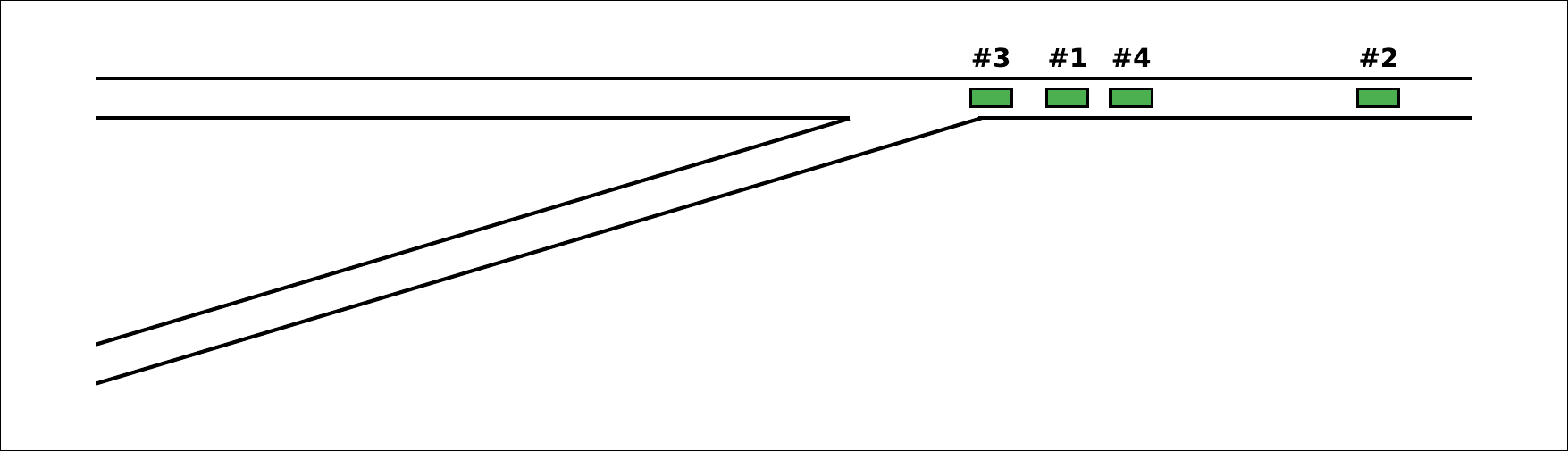}
  \caption{Case 2: timestep 15}
\end{subfigure}
\caption{Influence of adversarial faults introduced by the fault injection agent on fleet behavior. (a)-(b) Case 1: Safety compromise. (c)-(d) Case 2: Efficiency degradation.}
\label{fig:inj}
\vspace{-0.1in}
\end{figure}

A specific scenario is analyzed, in which vehicle 4 has perturbed observation when perceiving vehicle 2. In the first case, shown in Fig. \ref{fig:inj} (a) and (b), at timestep 9, the fault injection agent biases vehicle 4's observation of vehicle 2's position forward and its speed higher. This leads vehicle 4 to perceive a much larger time headway than reality, ultimately resulting in a collision between vehicle 2 and vehicle 4 at timestep 11. In the second case, depicted in Fig. \ref{fig:inj} (c) and (d), also at timestep 9, the fault injection agent acts conversely—biasing the observed position backward and the speed lower. Consequently, vehicle 4 decelerates to avoid a potential collision, triggering a chain reaction where vehicles 1 and 3 also brake successively. By timestep 15, the last vehicle, vehicle 3, finally passes the merging point, but is already significantly behind vehicle 2, indicating markedly reduced traffic efficiency.

The experiments in this section demonstrate that policies trained solely with random faults possess inadequate fault tolerance. In contrast, the fault injection agent can strategically select the most disruptive observation perturbation based on the specific scenario context, thereby effectively compromising either safety or efficiency.

\begin{table}[htbp]
\centering
\caption{Performance decline of methods trained with random faults: testing under random vs. adversarial faults}
\label{tab:inj}
\begin{tabular}{l l S[table-format=-1.2] S[table-format=+1.3, table-number-alignment=center] S[table-format=+2.2, table-number-alignment=center]}
\toprule
\textbf{Method} & \textbf{Fault Condition} & \textbf{Reward} & \textbf{Collision Rate} & \textbf{Timesteps} \\
\midrule
\multirow{3}{*}{\shortstack{Vanilla \\ MADDPG}} 
    & Random & {-1.26} & {21.2\%} & {17.39} \\
    & Adversarial & {-4.91} & {18.4\%} & {18.45} \\
    & $\Delta$ & {3.65}$\downarrow$ & {2.8\%} $\downarrow$ & {1.06}$\uparrow$ \\
\midrule
\multirow{3}{*}{\shortstack{OFT-MARL \\ w/o GRU}} 
    & Random & {-0.49} & {14.4\%} & {17.55} \\
    & Adversarial & {-5.56} & {10.6\%} & {18.90} \\
    & $\Delta$ & {5.07}$\downarrow$ & {3.8\%}$\downarrow$ & {1.35}$\uparrow$ \\
\midrule
\multirow{3}{*}{OFT-MARL} 
    & Random & {-0.14} & {12.9\%} & {17.54} \\
    & Adversarial & {-5.72} & {13.3\%} & {18.86} \\
    & $\Delta$ & {5.58}$\downarrow$ & {0.4\%} $\uparrow$ & {1.32} $\uparrow$ \\
\bottomrule
\end{tabular}
\begin{tabular}{p{\textwidth}}
\vspace{0.001in}
$\Delta$: Performance variation between two fault conditions \\
\end{tabular}
\vspace{-0.1in}

\footnotesize
\end{table}

\subsection{Training of Fault-tolerant Policies}
\label{sec: exp3}
\begin{figure}[htbp]
\includegraphics[width=\linewidth]{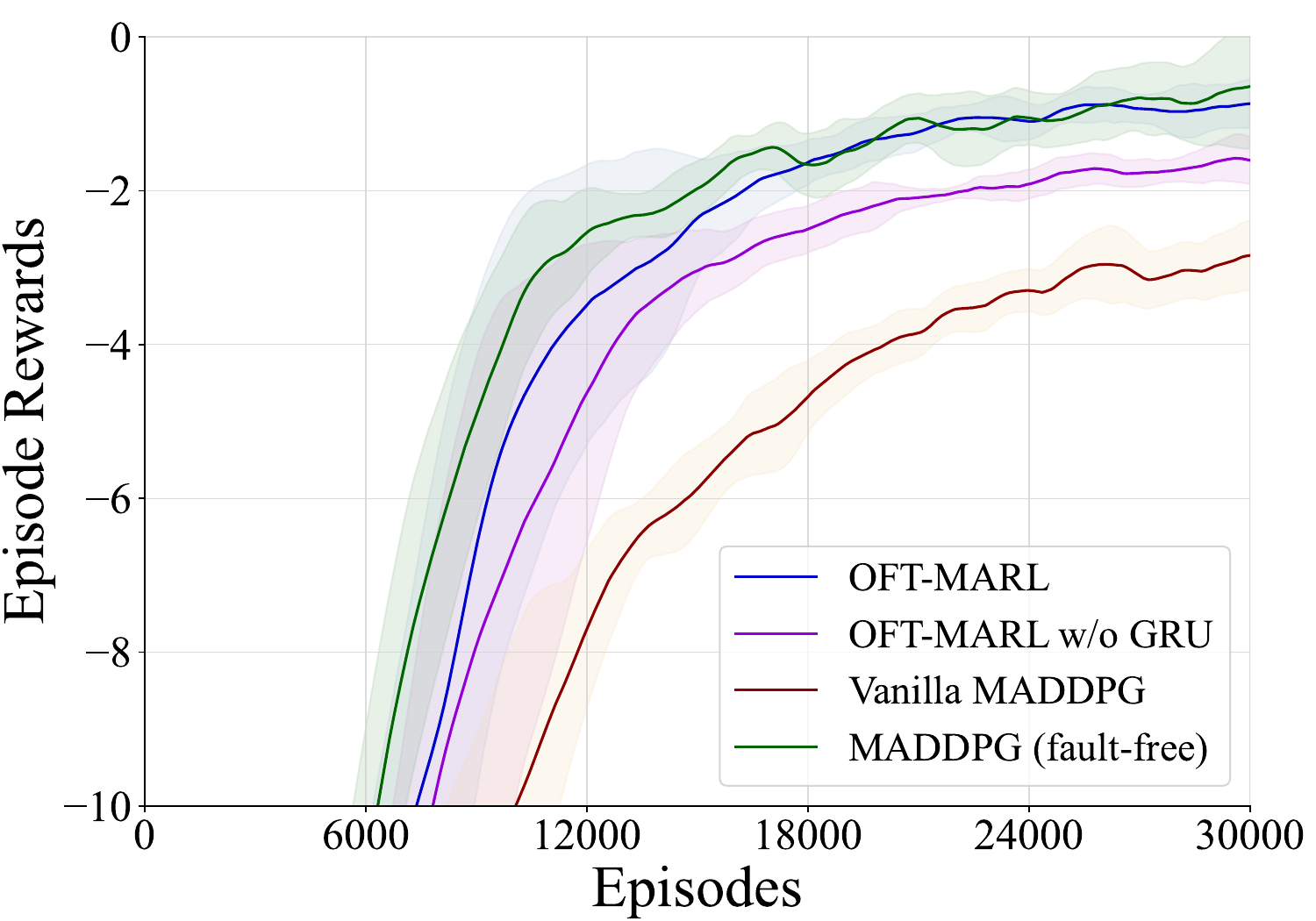}
\caption{Average reward curves for 3 methods trained with the fault injection agent and for MADDPG without considering faults. Error bars indicate 95\% confidence interval across 6 runs. }
\label{fig:t_curve}
\end{figure}

Having verified the efficacy of the fault injection agent, we employ it to generate adversarial faults during training, concurrently training both the agent and the vehicle policies. 
Under identical fault injection conditions, we compare the performance of different methods. As shown in Fig. \ref{fig:t_curve}, the standard MADDPG achieves the highest reward in the absence of faults. In contrast, vanilla MADDPG yields the poorest reward, as it fails to account for the impact of observation deviations on the accuracy of both critics and actors during training.
OFT-MARL w/o GRU builds upon vanilla MADDPG by adopting advanced training strategy of critics and actors according to Alg. \ref{alg:training_framework}. This modification leads to a substantial improvement in the performance of the trained vehicle policies. Furthermore, the proposed OFT-MARL introduces a temporal network that assists the policy network in detecting faults and estimating true observations from corrupted observations by leveraging historical information. This architecture further enhances the performance of OFT-MARL, enabling it to approach the performance of standard MADDPG under fault-free conditions.

As indicated by the test results in Table \ref{tab:t}, OFT-MARL outperforms other methods across all metrics, and its performance is close to that achieved in the fault-free scenario. These results demonstrate that the proposed method effectively mitigates the harmful effects of observation perturbations. Even when the fault injection agent deliberately introduces adversarial faults, OFT-MARL maintains performance levels comparable to those under fault-free conditions.

\begin{table}[htbp]
\centering
\caption{Performance comparison of methods trained with adversarial faults}
\label{tab:t}
\begin{tabular}{l S[table-format=-1.2] S[table-format=1.3] S[table-format=2.2]}
\toprule
\textbf{Method} & \textbf{Reward} & \textbf{Collision Rate} & \textbf{Timesteps} \\
\midrule
MADDPG (fault-free) & 0.30 & {11.3\%} & 17.52 \\
Vanilla MADDPG & -2.36 & {22.9\%} & 17.61 \\
OFT-MARL w/o GRU & -1.07 & {15.1\%} & 17.65 \\
OFT-MARL & -0.38 & {12.5\%} & 17.60 \\
\bottomrule
\end{tabular}
\end{table}

\begin{table*}[htbp]
\centering
\caption{Generalization performance evaluation under various fault injection conditions.}
\label{tab: generalization}
\begin{tabular}{l *{9}{c}}
\toprule
\multirow{2}{*}{Test Condition} & \multicolumn{3}{c}{OFT-MARL} & \multicolumn{3}{c}{OFT-MARL w/o GRU} & \multicolumn{3}{c}{Vanilla MADDPG} \\
\cmidrule(lr){2-4} \cmidrule(lr){5-7} \cmidrule(lr){8-10}
& Reward & Collision Rate & Timesteps & Reward & Collision Rate & Timesteps & Reward & Collision Rate & Timesteps \\
\midrule
Fault-free & \textbf{1.12} & {2.4\%} & 17.67 & 0.29 & {10.1\%} & 17.51 & -0.12 & {13.1\%} & 17.47 \\
Random fault & \textbf{-0.63} & {10.4\%} & 17.73 & -1.34 & {19.0\%} & 17.52 & -1.44 & {19.6\%} & 17.49 \\
OFT-MARL fault & \textbf{0.17} & {6.8\%} & 17.70 & -0.71 & {14.6\%} & 17.57 & -1.38 & {18.5\%} & 17.53 \\
OFT-MARL w/o GRU fault & -1.14 & {8.7\%} & 17.90 & \textbf{-0.88} & {13.2\%} & 17.67 & -2.12 & {15.9\%} & 17.83 \\
Vanilla MADDPG fault & \textbf{-0.32} & {8.2\%} & 17.74 & -1.33 & {17.8\%} & 17.58 & -1.07 & {17.6\%} & 17.51 \\
Retrained fault & \textbf{-1.63} & {10.1\%} & 17.97 & -2.32 & {15.2\%} & 17.93 & -5.21 & {21.6\%} & 18.36 \\
\bottomrule
\end{tabular}
\end{table*}

To evaluate the generalization capability of each method under diverse fault conditions, we select the best-performing policy from 6 random seeds for each approach and test it against multiple fault injection strategies. The fault injection methodologies encompassed: (1) fault-free conditions, (2) random faults, (3) faults generated by fault injection agents trained with the three vehicle policy training methods, and (4) faults generated by retrained fault injection agents with fixed vehicle policies.

As demonstrated in Table~\ref{tab: generalization}, the proposed OFT-MARL maintains fault-tolerant performance across various fault patterns, demonstrating generalization capability beyond its training-specific fault pattern. Notably, even when confronted with adversarially retrained fault injection policies targeting the fixed vehicle policy, the performance degradation remains marginal. The observed performance reduction is substantially smaller compared to the results presented in Table~\ref{tab:inj}. These findings collectively indicate that OFT-MARL exhibits considerable resilience against diverse fault patterns.

\subsection{Analysis of Vehicle Agent's Fault-Diagnosis Capability}
\label{sec: exp4}
To gain deeper insights into how the fault-tolerant mechanism within our vehicle agent contributes to enhancing fault tolerance, we conduct an independent analysis of the temporal network's fault-diagnosis outputs. First, we examine the accuracy of the temporal network in fault detection. Since the network outputs probability values and the dataset exhibits class imbalance, we binarize the predictions using a threshold of 0.5 to construct the confusion matrix. As shown in Table \ref{tab:matrix}, based on 69456 judgments collected over 1000 episodes, the network successfully identifies fault states 9296 times and correctly classifies normal states 59711 times. There are 412 instances where fault states are misclassified as normal, and 37 instances where normal states are falsely identified as faulty. This results in an accuracy of 99.3\%, a precision of 99.6\%, and a recall of 95.8\%.

Subsequently, we evaluate the accuracy of the temporal network in predicting true observations from corrupted observations. The results presented in Table~\ref{tab:recover} demonstrate that the network can effectively recover significant original observation errors to an acceptable level of prediction loss. For instance, considering the mean absolute error (MAE) of position estimation, the fault injection agent introduces an average position deviation of \SI{9.10}{\meter} between observed and true values. The temporal network reduces this error to \SI{3.36}{\meter}, corresponding to a 63.1\% error correction. Recovery rates also exceed 50\% among other metrics. The high accuracy in fault detection and the significant error reduction in observation estimation directly explain why the vehicle agent maintains high performance under faults. By providing reliable fault alerts and corrected state references, the temporal network empowers the policy network to make safer and more efficient decisions.

\begin{table}[h]
\centering
\caption{Confusion matrix of fault detection results}
\begin{tabular}{ccccc}
\toprule
\multicolumn{2}{c}{} & \multicolumn{2}{c}{Predicted} & \multirow{2}{*}{Total} \\
\cmidrule{3-4}
\multicolumn{2}{c}{} & Fault & Normal  \\
\midrule
\multirow{2}{*}{Actual} & Fault & 9296 & 412 & 9708 \\
                        & Normal & 37 & 59711 & 59748 \\
\midrule
\multicolumn{2}{c}{Total} & 9333 & 60123 & 69456 \\
\bottomrule
\label{tab:matrix}
\end{tabular}
\end{table}

\begin{table}[h]
\centering
\caption{Observation Prediction Recovery Percentage}
\begin{tabular}{cccccc}
\toprule
\multirow{2}{*}{} & \multicolumn{2}{c}{Position (\si{\meter})} & \multicolumn{2}{c}{Velocity (\si{\meter\per\second})} \\
\cmidrule(lr){2-3}\cmidrule(lr){4-5}
& MAE & MSE & MAE & MSE \\
\midrule
Original & 9.10 & 86.68 & 3.46 & 12.83 \\
\midrule
Prediction & 3.36 & 19.22 & 1.45 & 3.22 \\
\midrule
Recovery & 63.1\% & 77.8\% & 58.1\% & 74.9\% \\
\bottomrule
\label{tab:recover}
\end{tabular}
\end{table}

\begin{figure}[htbp]
\centering
\begin{subfigure}{\linewidth}
  \centering
  \includegraphics[width=\linewidth]{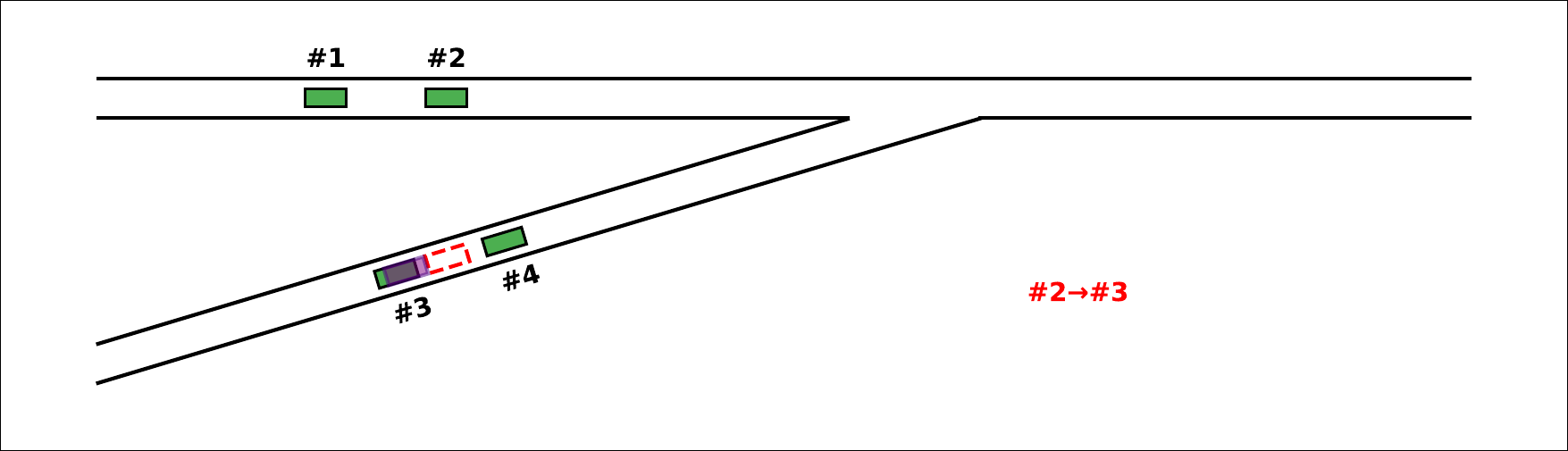}
  \caption{Case 1: timestep 4}
\end{subfigure}
\begin{subfigure}{\linewidth}
  \centering
  \includegraphics[width=\linewidth]{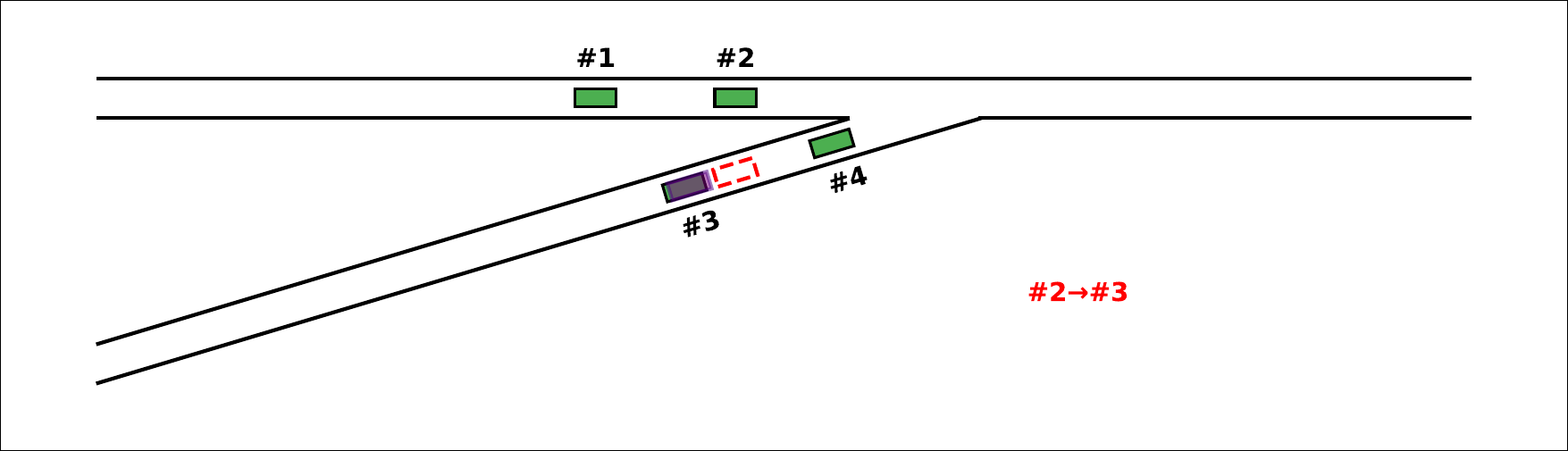}
  \caption{Case 1: timestep 9}
\end{subfigure}
\begin{subfigure}{\linewidth}
  \centering
  \includegraphics[width=\linewidth]{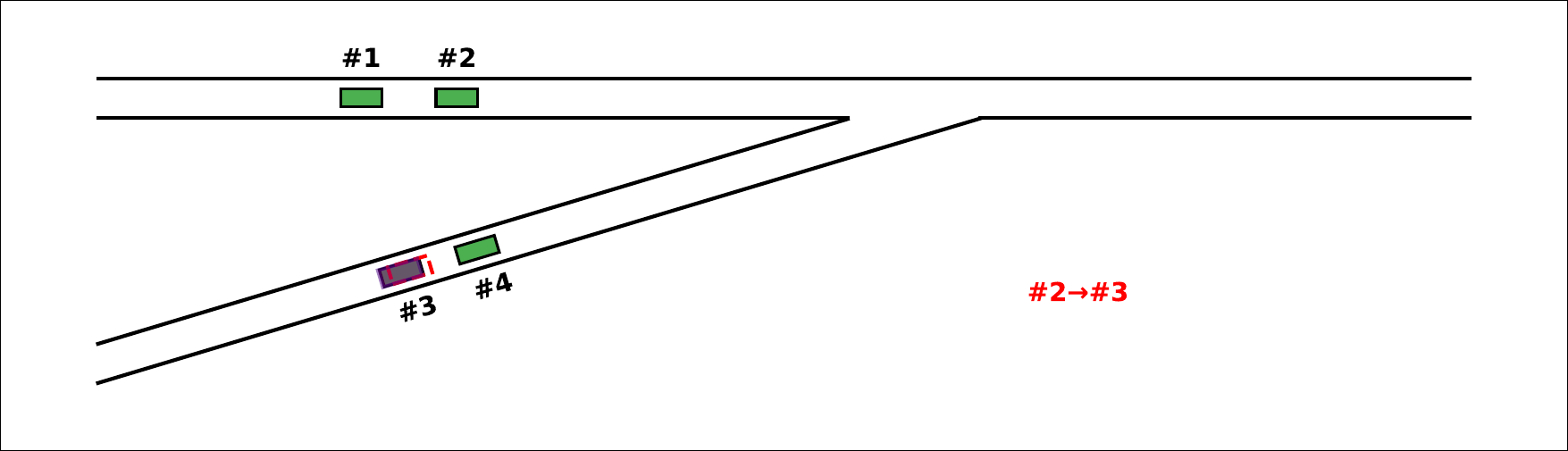}
  \caption{Case 2: timestep 4}
\end{subfigure}
\begin{subfigure}{\linewidth}
  \centering
  \includegraphics[width=\linewidth]{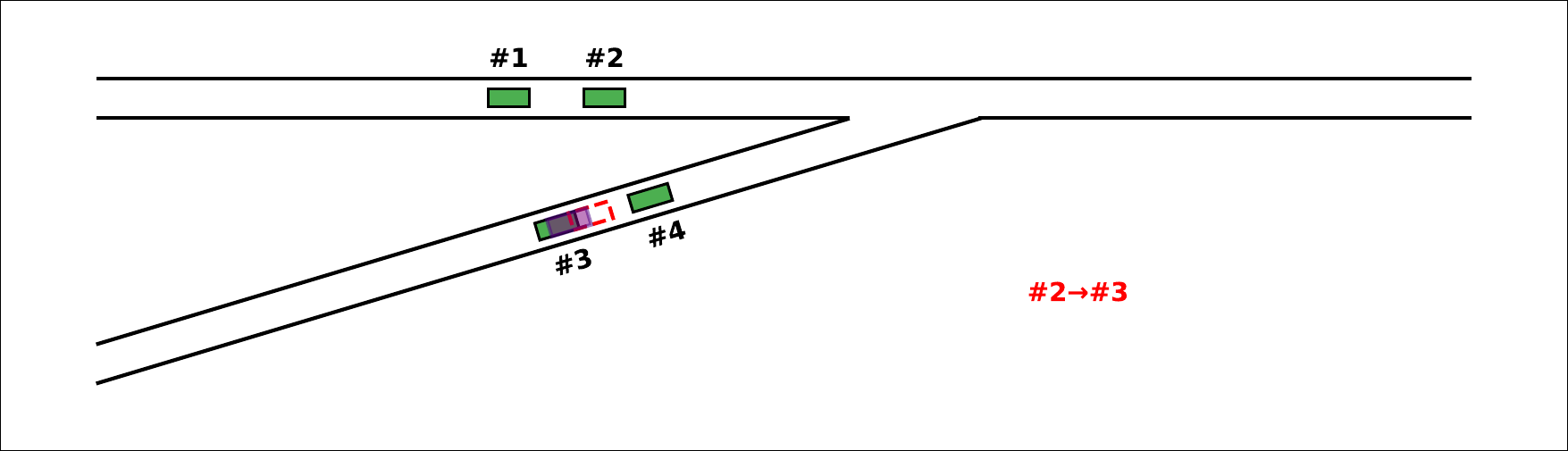}
  \caption{Case 2: timestep 7}
\end{subfigure}

\caption{Exhibition of the vehicle's fault-diagnosis capability in mitigating different fault patterns. (a)-(b) Case 1: Response to an abrupt position shift. (c)-(d) Case 2: Response to progressive deviations.}
\label{fig:tmp}
\end{figure}

To exhibit the performance of fault-tolerant mechanism under different fault patterns, we examine two representative cases as shown in Fig. \ref{fig:tmp}.

Case 1 investigates the network's response to abrupt state changes. At timestep 4, vehicle 2 observes a sudden forward position shift of \SI{6.0}{\meter} in vehicle 3. The network immediately detects this anomaly with high probability and generates a predicted position \SI{0.8}{\meter} ahead of the actual position, indicated by the purple rectangle.

Case 2 examines the network's capability in handling gradual state deviations. Starting from timestep 4, vehicle 2's observations of vehicle 3 exhibit progressively increasing errors: position deviation increases by \SI{1.0}{\meter} per timestep and velocity deviation by \SI{0.40}{\meter\per\second} per timestep. Although the observation prediction remains accurate at timestep 4, the network fails to initially identify the fault due to minor deviations. However, as vehicle 3 continues to accelerate in vehicle 2's view, which contradicts the normal driving pattern, the temporal network gradually increases its fault probability assessment. At timestep 7, vehicle 2 successfully identifies the fault. The observed position and velocity contain a \SI{4.0}{\meter} and \SI{1.60}{\meter\per\second} forward error respectively. The network's predictions substantially correct these deviations, reducing the position error to \SI{1.4}{\meter} and the velocity error to \SI{0.36}{\meter\per\second}.

In both scenarios shown in panels (b) and (d), without the diagnostic capability, vehicle 2 would misinterpret vehicle 3's behavior as aggressive positioning and potentially initiate unnecessary deceleration. This could either cause collisions with actually positioned vehicles or significantly increase traffic delay. Equipped with our fault-tolerant design, vehicle 2 correctly maintains speed and safely precedes vehicle 3 through the conflict point. 

\section{Conclusion}
\label{sec:Conclusion}
This paper highlights the severe impact of observation perturbations on the safety and efficiency of MARL-based cooperative driving systems. To address this, we introduce a fault-tolerant MARL framework integrating an adaptive fault injection agent and a fault-tolerant vehicle agent. The fault injection agent actively generates challenging perturbations during training, exposing agents to critical faults and enhancing fault tolerance. The vehicle agent is equipped with the capability to detect faults and reconstruct credible observation estimates by leveraging spatio-temporal vehicle data. Experimental results in highway on-ramp merging scenarios confirm that our approach significantly outperforms vanilla MADDPG and ablated variants under both random and adversarial fault conditions. It nearly matches the performance of fault-free systems while maintaining strong generalization across unseen fault patterns. The temporal network achieves high fault detection accuracy and substantially corrects observation errors.

In future research, the applicability of the method across diverse scenarios can be investigated. Moreover, it can be integrated with safety verification modules to ensure the security of vehicle policies. Additionally, more data credibility assessment methods, such as incorporating driving habits factors, could be fused into the vehicle agent to enhance fault detection and observation prediction capabilities for heterogeneous vehicles.

\bibliographystyle{IEEEtran}
\bibliography{liter}

\end{document}